\documentclass[10pt,twocolumn,letterpaper]{article}

\usepackage{cvpr}
\usepackage{times}
\usepackage{epsfig}
\usepackage{graphicx}
\usepackage{amsmath}
\usepackage{amssymb}

\usepackage{subfig}

\usepackage[breaklinks=true,bookmarks=false]{hyperref}

\cvprfinalcopy 

\def\cvprPaperID{****} 

\begin{document}

\title{Unsupervised learning of object semantic parts from internal states of CNNs by population encoding}

\author{Jianyu Wang$^1$, Zhishuai Zhang$^{2}$, Cihang Xie$^{2}$, Vittal Premachandran$^2$, Alan Yuille$^{1,2}$ \\
$^1$University of California Los Angeles, $^2$Johns Hopkins University \\
\texttt{wjyouch@ucla.edu, \{zhshuai.zhang, cihangxie306\}@gmail.com}, \\ 
\texttt{\{vittalp, alan.yuille\}@jhu.edu}
}

\maketitle

\begin{abstract}
We address the key question of how object part representations can be found from the internal states of CNNs that are trained for high-level tasks, such as object classification. This work provides a new unsupervised method to learn semantic parts and gives new understanding of the internal representations of CNNs. Our technique is based on the hypothesis that semantic parts are represented by populations of neurons rather than by single filters. We propose a clustering technique to extract part representations, which we call Visual Concepts. We show that visual concepts are semantically coherent in that they represent semantic parts, and visually coherent in that corresponding image patches appear very similar. Also, visual concepts provide full spatial coverage of the parts of an object, rather than a few sparse parts as is typically found in keypoint annotations. Furthermore, We treat each visual concept as part detector and evaluate it for keypoint detection using the PASCAL3D+ dataset, and for part detection using our newly annotated ImageNetPart dataset. The experiments demonstrate that visual concepts can be used to detect parts. We also show that some visual concepts respond to several semantic parts, provided these parts are visually similar. Note that our ImageNetPart dataset gives rich part annotations which cover the whole object, making it useful for other part-related applications.
\end{abstract}

\section{Introduction}
Deep Convolutional Neural Networks (CNNs) are very successful at performing  visual tasks such as image classification \cite{KrizhevskySH12,SimonyanZ14a}, object detection \cite{GirshickDDM14}, and semantic segmentation \cite{longfully,ChenPKMY14}. Despite these successes we lack understanding of why CNNs work so well. For example, when CNNs are trained to detect objects it seems likely that the object semantic parts are represented by the internal activity of the filters/neurons. But what is the form of these internal representation? Are semantic parts represented by individual neurons \cite{ZhouKLOT14} or perhaps by populations of activity \cite{georgopoulos1986neuronal}? Understanding the nature of these representations not only makes it easier to understand deep networks, and hence may suggest ways to improve them, but also serves as an unsupervised method for learning object semantic parts. Note, we use the term ``semantic parts" to mean object parts, like wheels and windows of cars, which are defined in terms of the three-dimensional object.

We address this problem by studying the internal activity of CNNs when looking at six classes of objects (car, aeroplane, bike, motorbike, bus, and train). Our hypothesis is that object semantic parts are represented by populations of filter/neural activity which we can find by clustering algorithms. We call each cluster a ``visual concept". Although these visual concepts are defined in terms of the feature activity of the CNN they also correspond to a set of image patches (those patches whose feature activity lie within the cluster). We show visually that visual concepts (i.e. the image patches associated with them) correspond well to parts of objects (e.g.,  the wheels and windows of cars). These clusters are also visually tight, in the sense that image patches corresponding to the same visual concept tend to look very similar. Moreover, the visual concepts give a fairly dense description of each object, in the sense that for most parts of the object we can find a visual concept that looks visually similar.

We proceed to understand and quantify the correspondence between the visual concepts and the semantic parts of the objects. The correspondence is non-trivial because the scale of the visual concepts and semantic parts may differ, hence several visual concepts may correspond to different, or overlapping, regions of the same semantic part. Conversely, some semantic parts may look visually similar (e.g.,  the bike wheel center and the bike chain ring). In general, we expect the correspondence between visual concepts and semantic parts to be many-to-many.

To study this correspondence we convert each visual concept into a part detector and evaluate them in terms of part detection and localization. We first compute the distance between the center of the visual concept and the feature response/activity to an input image patch. If the distance is below a threshold we treat this as a detection of that visual concept. We then evaluate the visual concepts for detecting keypoints in the PASCAL3D+ dataset \cite{xiang2014beyond}. This gives the promising result that for each keypoint there is typically a visual concept that detects the keypoint fairly well, as measured by average precision (AP), and compared with detection using single filters or a supervised baseline method. But this study is limited because the keypoints are very sparse and only cover a small portion of each object, so many visual concepts do not correspond to them.

To overcome this difficulty we create a new dataset, ImageNetPart, by annotating the six object classes (car, aeroplane, bike, motorbike, bus, and train) in PASCAL3D+ \cite{xiang2014beyond} with dense semantic parts and also determining the background regions. We use the ImageNet images provided in PASCAL3D+ dataset. This is a total of roughly 10,000 images and requires roughly 6 times as many labels as the keypoints. As before, we treat the single visual concepts as part detectors and show that for most semantic parts there is a visual concept that detects it fairly well (as measured by AP). But this only tells us what a subset of visual concepts are doing, because the number of visual concepts is much larger than the number of semantic parts (by a factor of 4 or more). By further analysis we show that the remaining visual concepts fall into three classes: (i) those that detect several semantic parts (two to four) which are visually similar, (ii) those that detect background regions of images (e.g., sky for aeroplanes, railway tracks for trains), and (iii) a few which have no obvious correspondence (perhaps due to limitations of the clustering method and CNN feature). We obtain better APs in evaluation when taking into account the fact that visual concepts may respond to a small subset of semantic parts.

We argue that our work gives some understanding of CNNs and, in particular, of how they encode internal representations of objects by visual concepts, encoded by populations of filter/neuron activity. From another perspective, our approach can be thought of as an unsupervised way to learn part detectors of objects.

\section{Related Work}
\label{sec:related}

Discovering mid-level patches \cite{singh2012unsupervised,juneja2013blocks} in an unsupervised manner has been studied before CNNs. More recently, \cite{LiLSH15} used CNN features with association rule algorithm to discover mid-level visual elements, and evaluated them on object and scene classification tasks. But their method used CNN only as a feature extractor without exploring how the CNN filters of intermediate layers capture mid-level visual concepts. \cite{XiaoXYZPZ15} and \cite{SimonR15} used activations from  intermediate layers of the CNN to find object parts, which they then used for fine-grained object classification. But they assumed that a single filter from an intermediate layer can detect parts. Our work has some similarity to \cite{ZhouKLOT14}, which looked at Scene-CNNs (trained for scene classification), and claimed that object detectors emerge from Scene-CNNs. Their ``object detectors" are single filters/neurons which they evaluated using AMT. By comparison, we extract visual concepts at multiple levels of CNN in an unsupervised manner, and evaluate them as part detector automatically.

There is some neuroscience evidence that the brain represents concepts by population activity \cite{georgopoulos1986neuronal}, but there have been few studies of this in CNNs. One work \cite{AgrawalGM14} argued for a population code by training linear SVMs on features from Alexnet and showed that several filters were needed to get good performance on object classification. But, unlike us, their approach was fully supervised. They also addressed objects instead of object parts. There has been other supervised methods for object part detection \cite{chendetect,long2014convnets} which show good performance.

Several researchers have visualized deep networks. \cite{erhan_visualizing,SimonyanVZ13} used backpropogation-based techniques to generate images by maximizing certain object class score, enabling them to visualize the notion of an object class as captured by CNN. And \cite{MahendranV15} showed that incorporating strong natural image priors to the optimization procedure produces images that look more realistic. \cite{zeiler2014visualizing} proposed a visualization strategy that mapped intermediate features activations back to the image space using Deconvolution Networks \cite{zeiler2011adaptive}.


\begin{figure}[t!]
\centering
\includegraphics[width=1\linewidth]{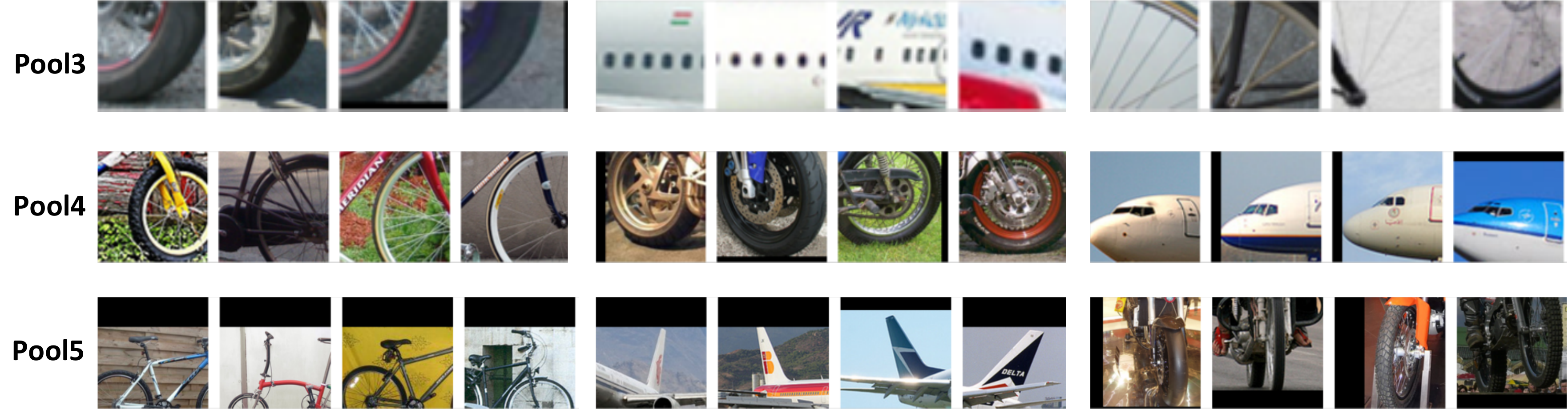}
\caption{\small This figure shows example visual concepts on different object categories, for layers \texttt{pool3}, \texttt{pool4} and \texttt{pool5}. Each row visualizes three visual concepts with four example patches, which are randomly selected from a pool of well-matched patches. These visual concepts are visually and semantically tight. We can easily identify the semantic meaning and parent object class.}
\vspace{-0.5cm}
\label{fig:concept_viz}
\end{figure}

\begin{figure}[t!]
\centering
\subfloat[\texttt{pool3} visual concept]
{\includegraphics[width=.48\linewidth]{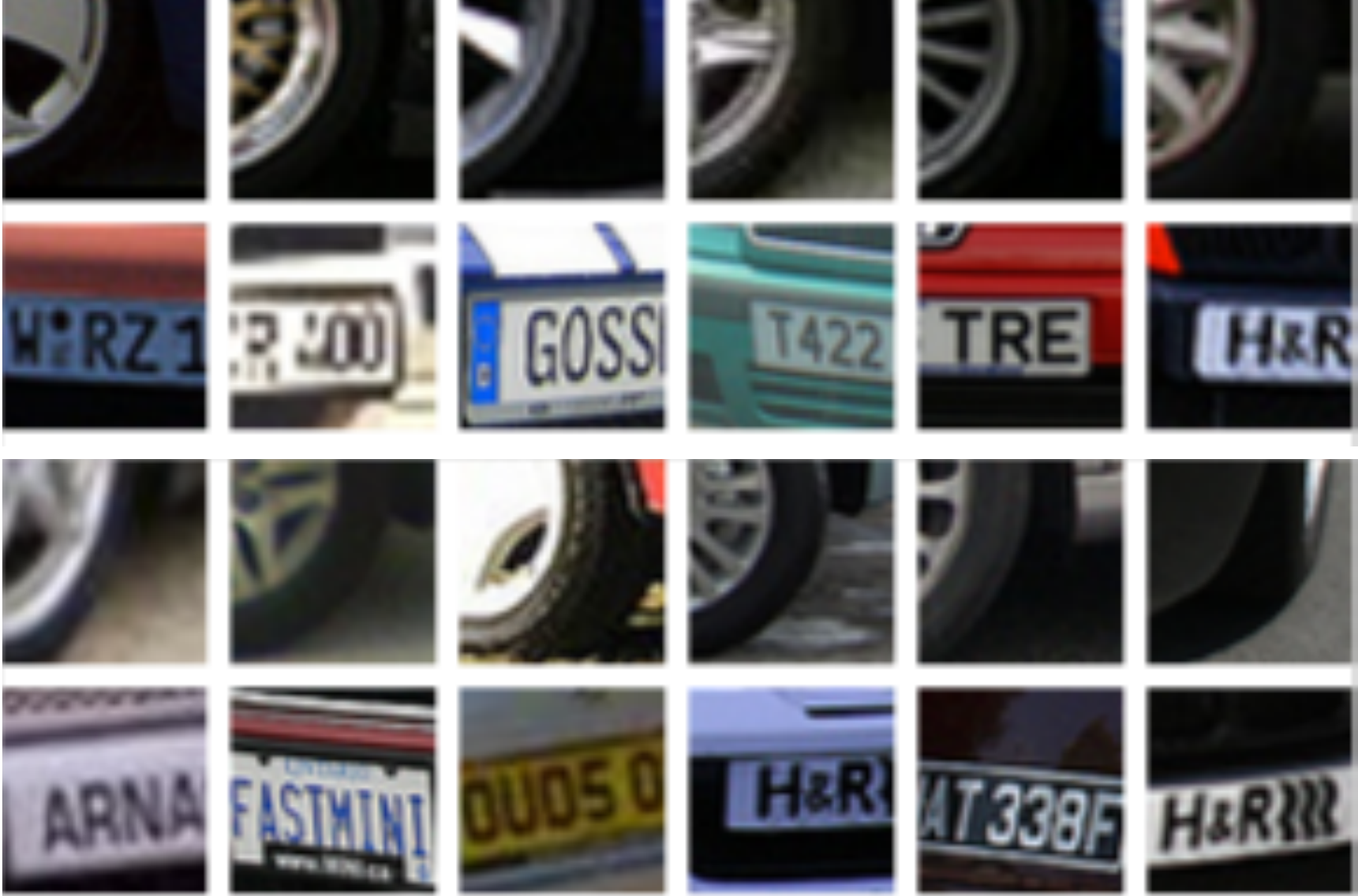}\label{fig:pool3_vis_cluster}}
\hfill
\subfloat[\texttt{pool3} single filter]
{\includegraphics[width=.48\linewidth]{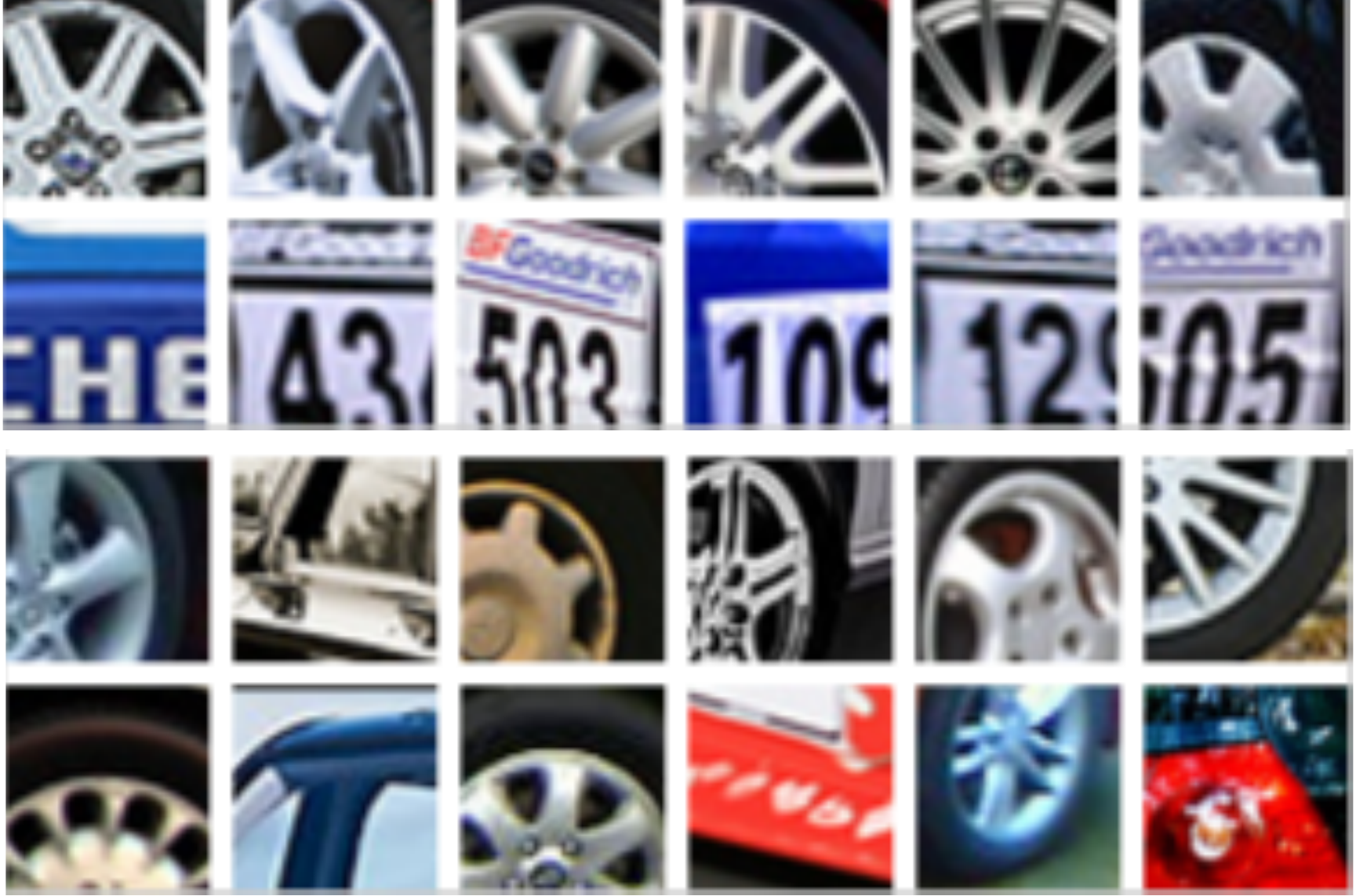}\label{fig:pool3_vis_single}}
\vspace{-0.3cm}

\subfloat[\texttt{pool4} visual concept]
{\includegraphics[width=.48\linewidth]{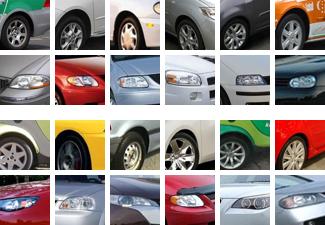}\label{fig:pool4_vis_cluster}}
\hfill
\subfloat[\texttt{pool4} single filter]
{\includegraphics[width=.48\linewidth]{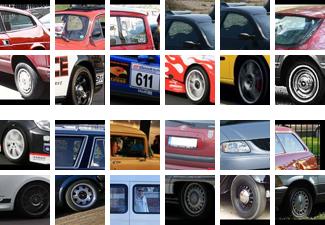}\label{fig:pool4_vis_single}}

\subfloat[\texttt{pool5} visual concept]
{\includegraphics[width=.48\linewidth]{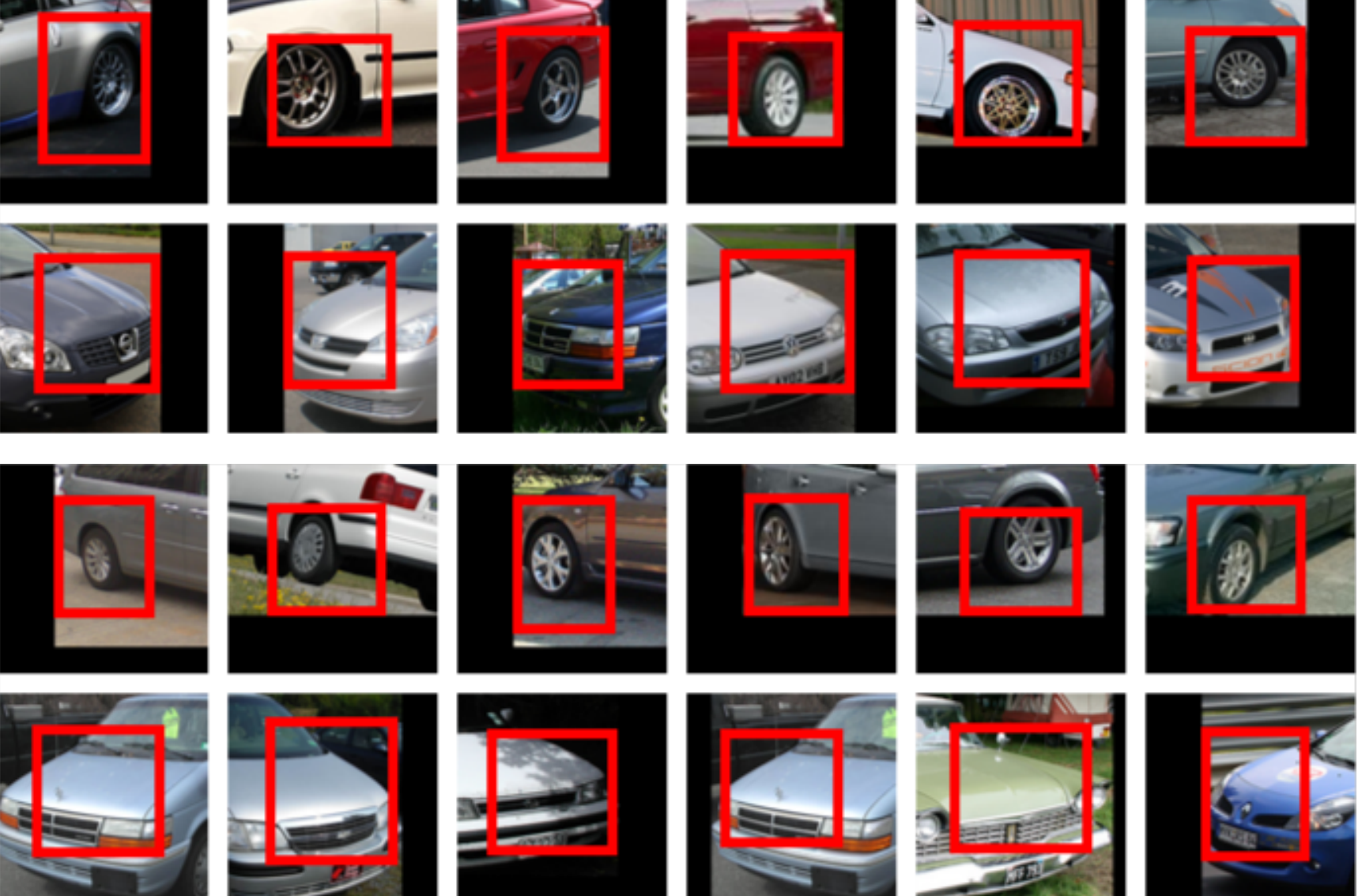}\label{fig:pool5_vis_cluster}}
\hfill
\subfloat[\texttt{pool5} single filter]
{\includegraphics[width=.48\linewidth]{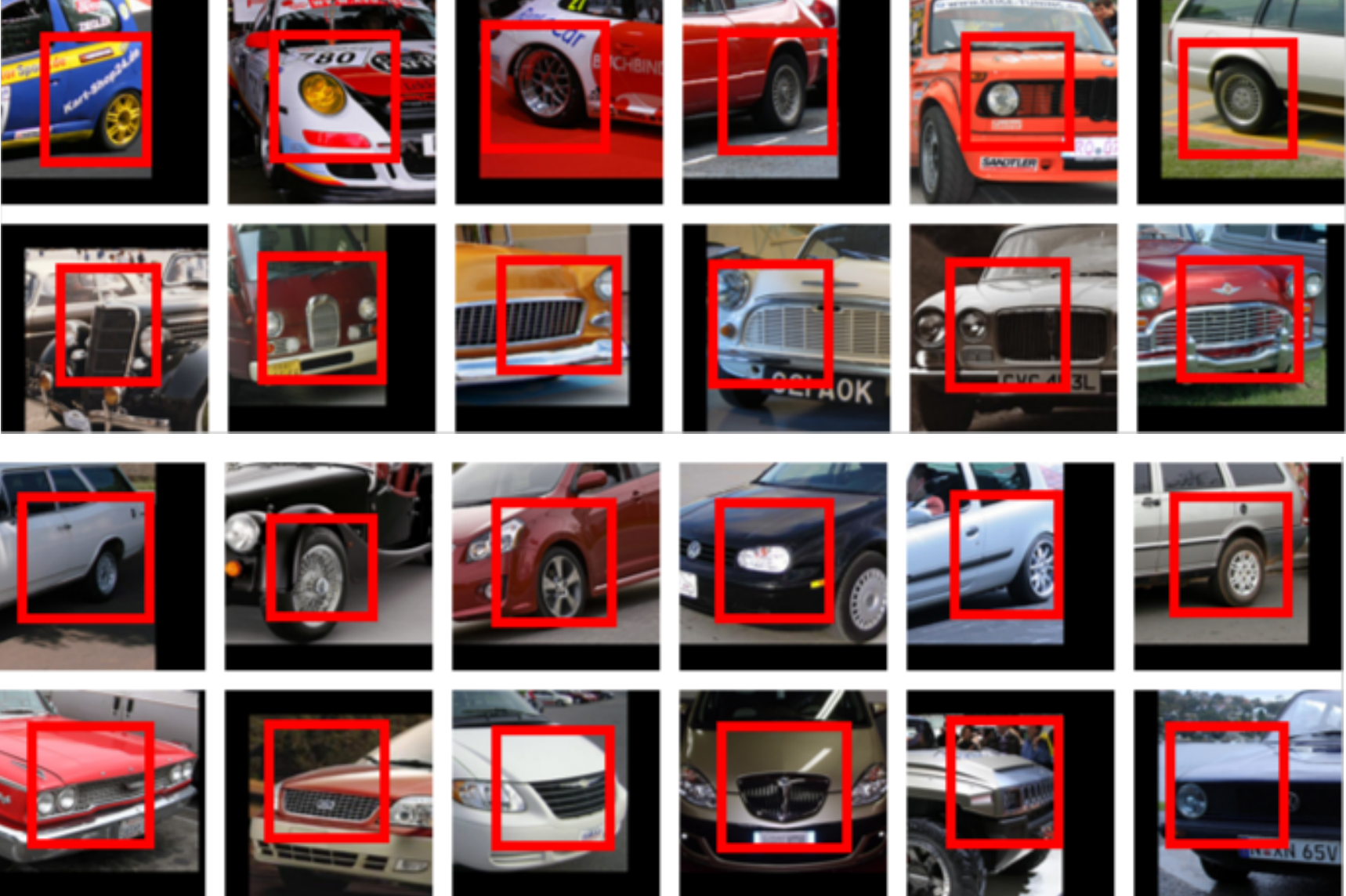}\label{fig:pool5_vis_single}}
\caption{\small Visual comparison between visual concepts and single filters for layer \texttt{pool3}, \texttt{pool4} and \texttt{pool5}. In subfigure (a), we visualize two randomly selected visual concepts, where the top two rows correspond to the best 6 patches, and the bottom two rows correspond to 6 randomly selected patches from the top 500 well-matched patches. In subfigure (b), we visualize two randomly selected single filters in the same manner.}
\vspace{-0.3cm}
\label{fig:viz}
\end{figure}

\section{Learning Visual Concepts by Clustering}
\label{sec:population}
This section describes our method for finding visual concepts in object-CNNs. We start with a CNN pre-trained for ImageNet object classification. We apply this CNN to a set of images from a specific object class, such as cars. Then we cluster the features at each layer of CNN (ignoring the spatial locations of the features) using  K-means++ \cite{arthur2007k},  yielding a dictionary of visual concepts at each layer. We will understand the CNNs internal representations better by restricting the input to objects from the same class, rather than objects from many classes.

More precisely, for an image $\mathbf{x} \in \mathbb{R}^{W \times H \times 3}$ with spatial resolution $W \times H$, an intermediate layer, $\ell$, produces a response, $\mathbf{f}_\ell \in \mathbb{R}^{W_\ell \times H_\ell \times N_\ell}$,  where $W_\ell \times H_\ell$ is the spatial resolution of the intermediate layer $\ell$'s response and $N_\ell$ is the number of filters on that layer. We randomly sample a number of feature population responses, $\mathbf{p}^{i,j}_\ell \in \mathbb{R}^{N_\ell}$, over the spatial grid $(i,j)$ of the layer's response $\mathbf{f}_\ell$. Each $\mathbf{p}^{i,j}_\ell$ corresponds to a patch from the original image with the theoretical receptive field, which can be computed for any spatial grid $(i, j)$ and any intermediate layer $\ell$. We extract the population responses, $\mathbf{p}^{i,j}_\ell$, from all images in the training dataset. Next we cluster the population responses, $\mathbf{p}^{i,j}_\ell$, using the K-means++ algorithm. At each level of the CNN, this gives a dictionary of visual concepts. The intuition is that this clustering will identify image patterns that frequently occur in images of cars and hence which may correspond to parts such as car wheels or license plates. An important issue is to determine the number of visual concepts $K$. We adopt a two-step strategy where we first set $K$ in kmeans++ to be the number of filters in the layer, and then we reduce the dictionary size by a greedy cluster merging algorithm. We analyze the effect of $K$ in the diagnostic experiment.

\subsection{Visualization of Visual Concepts}
\label{sec:viz}

We can visualize the visual concepts by observing the image patches which are assigned to each cluster by K-means++. Our first finding is that the image patches for each visual concept appear to roughly correspond to semantic parts of the object classes with larger parts at higher pooling levels, see figure \ref{fig:concept_viz}. Most of the visual concepts are tight in the sense that the image patches corresponding to them look visually similar. But demonstrating this in a conference paper is difficult for two reasons (i) it is impractical to show all the image patches for each visual concept, and (ii) we also cannot show all the visual concepts. To address the second difficulty we show all the visual concepts of cars in the supplementary material with comparisons to the individual filters.

\begin{figure}[t]
\centering
\captionsetup{captionskip=-8pt}
\subfloat[\texttt{pool4} visual concept]{
\begin{minipage}[c][3.5cm][t]{.111\linewidth}
  \centering
  \includegraphics[width=1cm,height=1cm]{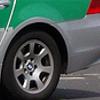}
  \includegraphics[width=1cm,height=1cm]{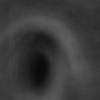}
  \includegraphics[width=1cm,height=1cm]{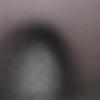}
\end{minipage}
\begin{minipage}[c][3.5cm][t]{.111\linewidth}
  \centering
  \includegraphics[width=1cm,height=1cm]{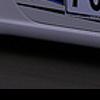}
  \includegraphics[width=1cm,height=1cm]{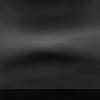}
  \includegraphics[width=1cm,height=1cm]{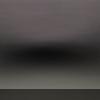}
\end{minipage}
\begin{minipage}[c][3.5cm][t]{.111\linewidth}
  \centering
  \includegraphics[width=1cm,height=1cm]{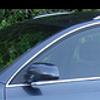}
  \includegraphics[width=1cm,height=1cm]{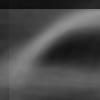}
  \includegraphics[width=1cm,height=1cm]{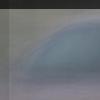}
\end{minipage}
\begin{minipage}[c][3.5cm][t]{.111\linewidth}
  \centering
  \includegraphics[width=1cm,height=1cm]{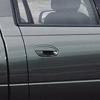}
  \includegraphics[width=1cm,height=1cm]{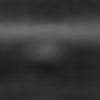}
  \includegraphics[width=1cm,height=1cm]{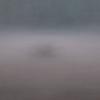}
\end{minipage}
}
\hfill
\subfloat[\texttt{pool4} single filter]
{
\begin{minipage}[c][3.5cm][t]{.111\linewidth}
  \centering
  \includegraphics[width=1cm,height=1cm]{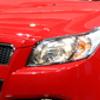}
  \includegraphics[width=1cm,height=1cm]{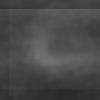}
  \includegraphics[width=1cm,height=1cm]{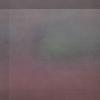}
\end{minipage}
\begin{minipage}[c][3.5cm][t]{.111\linewidth}
  \centering
  \includegraphics[width=1cm,height=1cm]{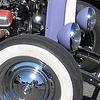}
  \includegraphics[width=1cm,height=1cm]{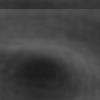}
  \includegraphics[width=1cm,height=1cm]{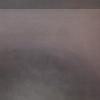}
\end{minipage}
\begin{minipage}[c][3.5cm][t]{.111\linewidth}
  \centering
  \includegraphics[width=1cm,height=1cm]{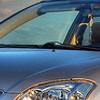}
  \includegraphics[width=1cm,height=1cm]{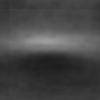}
  \includegraphics[width=1cm,height=1cm]{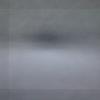}
\end{minipage}
\begin{minipage}[c][3.5cm][t]{.11\linewidth}
  \centering
  \includegraphics[width=1cm,height=1cm]{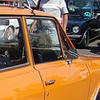}
  \includegraphics[width=1cm,height=1cm]{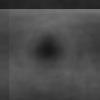}
  \includegraphics[width=1cm,height=1cm]{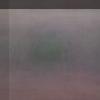}
\end{minipage}
}
\caption{\small This figure shows four visual concept examples (left four) and four single filter examples (right four). The top row contains example image patches, and the bottom two rows contain average edge maps and intensity maps respectively obtained using top 500 patches. Observe that the means of the visual concepts are sharper than the means of the single filters for both average edge maps and intensity maps, showing visual concepts capture patterns more tightly than the single filters.}
\label{fig:avg}
\end{figure}

\begin{figure}[t!]
\centering
\includegraphics[trim={2.5cm 4cm 5cm 6.5cm},clip,width=0.23\textwidth]{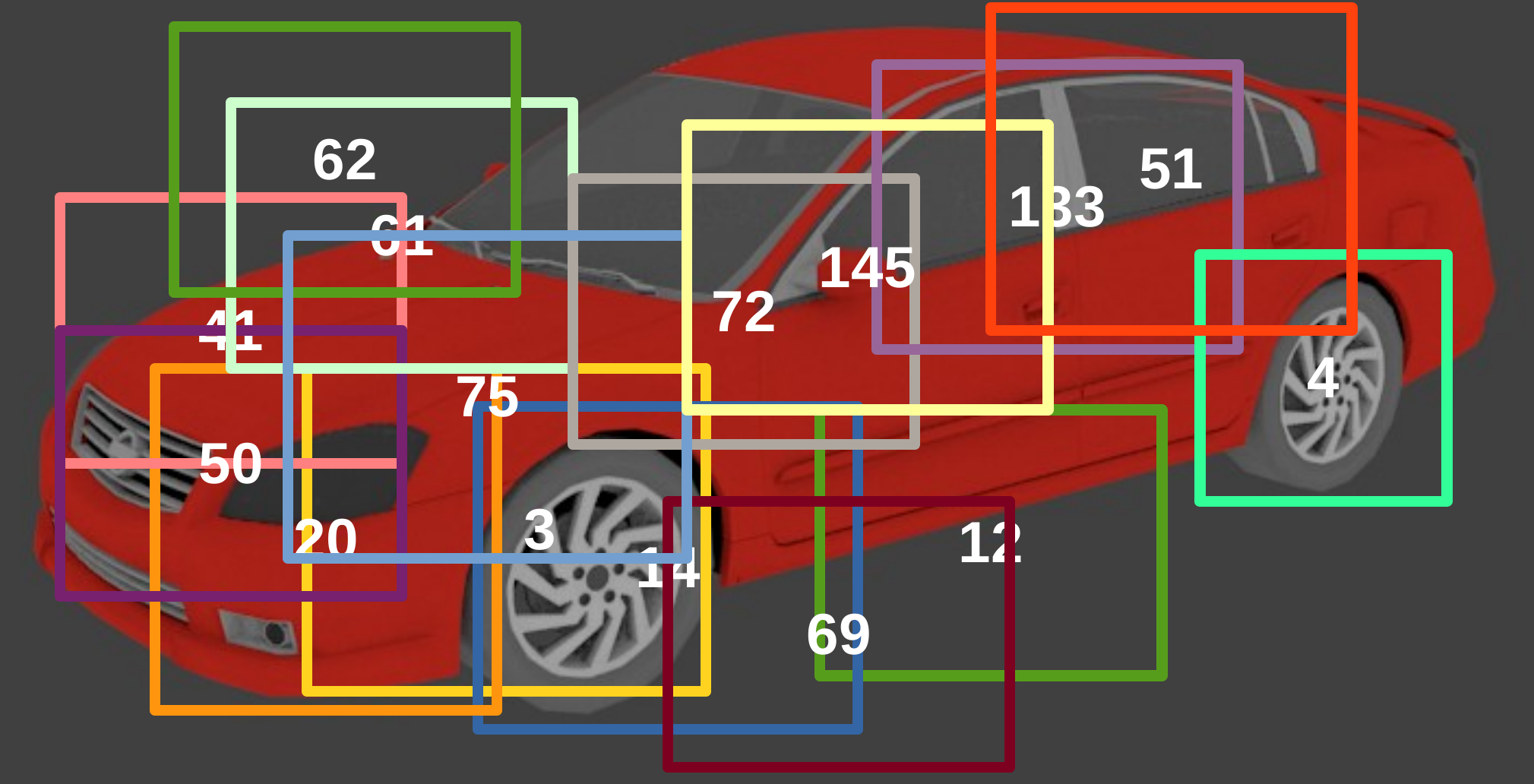}
\hfill
\includegraphics[trim={0 3cm 0 3.5cm},clip,width=0.23\textwidth]{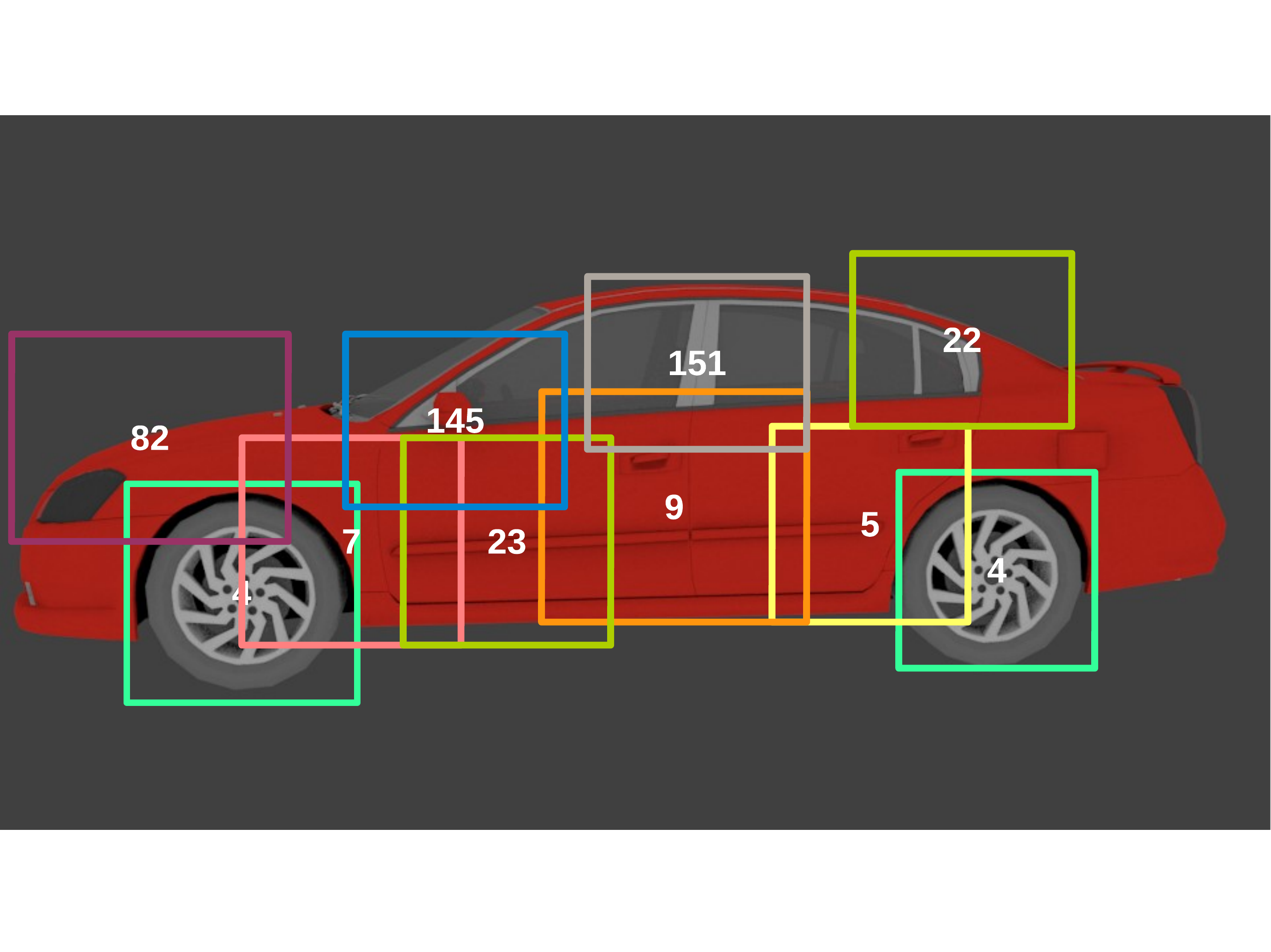}
\caption{\small The figure shows example cars rendered by 3D model. We overlay the image with bounding boxes that are denoted by the visual concept number at the center, which would be the best match at that location. We see that the learned visual concepts for cars can cover most of the car region.}
\label{fig:coverage}
\vspace{-0.5cm}
\end{figure}

We address the first difficulty in two ways. Firstly, in figure \ref{fig:viz} we show a small number of the best image patches for each visual concept (i.e. the best 6 whose feature vectors are closest to the visual concept center) and a random sample of the other patches (e.g., the random 6 from best 500). For comparison, we show similar results for individual filters (chosen from the small fraction of filters that appear to correspond to semantic parts). We observe that best patches and the random samples for visual concepts are often very visually similar, but this is less true for the individual filters. Secondly, for each visual concept we compute (a) the mean of the edge maps of the best 500 image patches using HED \cite{xie15hed} and (b) the average of the (color) intensity of the best 500 patches. Examples of these means are given in figure \ref{fig:avg}. These both show that most visual concepts are visually tight. Furthermore figure \ref{fig:coverage} is a schematic showing that visual concepts give a spatially complete description of object parts and cover most of the object. The supplementary material extends figures \ref{fig:viz} to all the visual concepts at \texttt{pool4} and all the filters at that level. Roughly speaking, around seventy percent of the visual concepts are tight but this is only true for roughly thirty percent of the filters.

Finally we mention an artifact that arises when we visualize the visual concepts and has implications when evaluating them as part detectors. We observed that the visual concepts appeared less tight at the highest levels of the hierarchy (e.g. pool5). We conjecture that one reason for this is that the ``effective receptive field” of the neurons is not the same as their theoretical receptive fields. Intuitively, the effective receptive field captures the region of pixels which are most responsible for the response of visual concept or single filter. We obtain the effective receptive field by applying the deconvolutional operation in (Zeiler and Fergus 2014), and then threshing the deconv responses. We find out that, for visual concepts at pool5, the image patches look more similar within the effective receptive field (red bounding box) than looking at the entire patch (which is very large).

\section{Visual Concepts as Part Detectors}
\label{sec:quant}
Our visualization studies suggest that the visual concepts roughly correspond to semantic parts of objects and we now  quantify this by evaluating them as part detectors.

We treat a visual concept as a part detector as follows. We compute the distance in the CNN feature space between the center of visual concept and the filter response to an image patch containing, or not containing, the part. If this distance is below a threshold we treat this as a detection for that visual concept. We also use non-maximal suppression (NMS) to suppress duplicate detections. If the detected patch center is sufficiently close (e.g., within 56 pixels) to the (keypoint or semantic part) groundtruth position, we treat it as a true positive. In this way, for each visual concept and semantic part we compute the number of true positives and false positives as a function of the threshold. By varying this threshold we obtain precision recall curves, and calculate the average precision (AP).

To evaluate keypoint detection in PASCAL3D+, or semantic part detection in ImageNetPart, we select the visual concept which has highest AP. This ``best visual concept" evaluation strategy is problematic if the visual concept is good at detecting several keypoints or semantic parts. Later on we modify our evaluation method to deal with this. By visualization we find that visual concepts from \texttt{pool4} layer corresponds best to the keypoint annotations in PASCAL3D+ and the semantic part annotations in ImageNetPart, so we report them in this paper. The results for \texttt{pool5} are fairly similar, while the results for \texttt{pool3} are worse presumably because their visual concepts are too small.

\begin{figure}[t]
\includegraphics[width=0.19\textwidth]{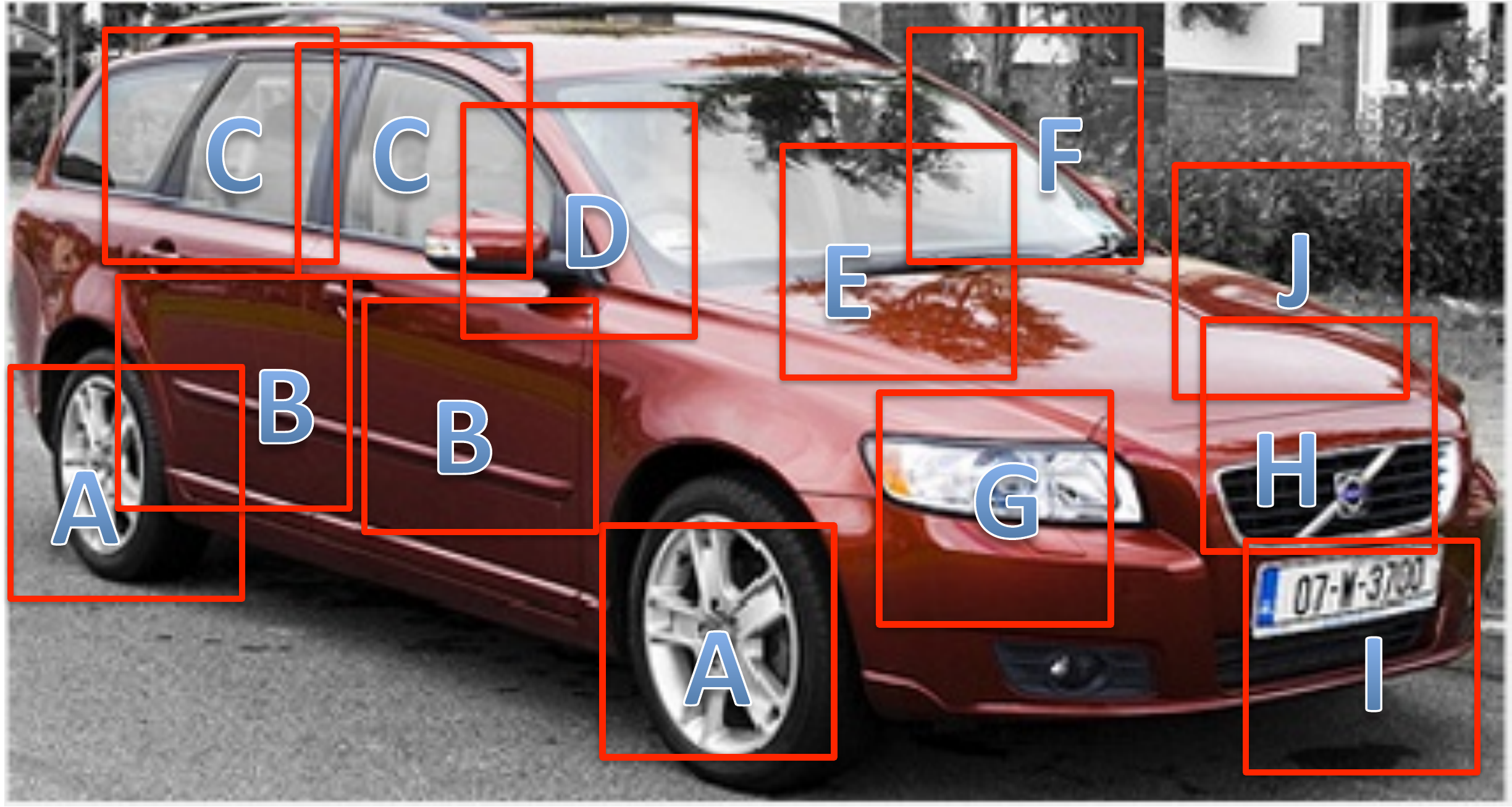}
\includegraphics[trim={0cm 0.8cm 0cm 0cm},clip,width=0.28\textwidth]{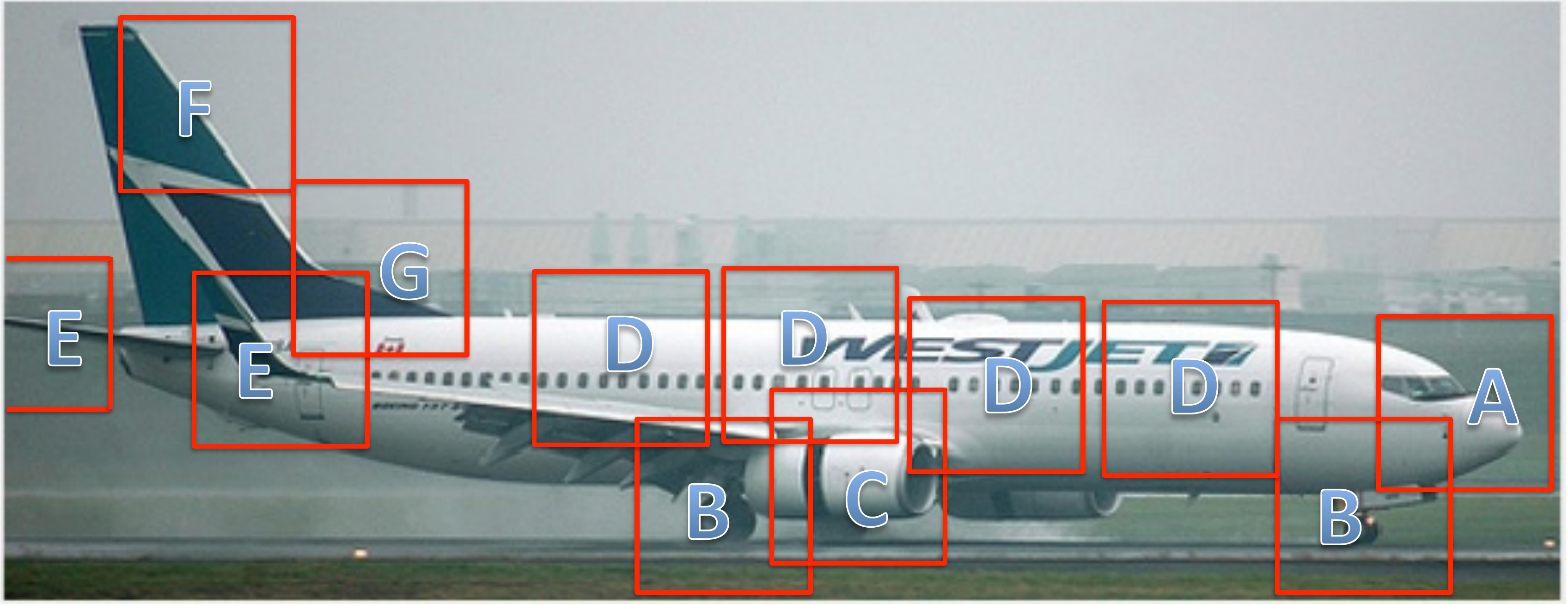}
\caption{\small The figure illustrates several semantic part annotations for cars (left) and aeroplanes (right) from a particular viewpoint. We have defined in total 35 semantic parts for cars and 17 for aeroplanes. Car -- A: wheels, B: side body, C: side window, D: side mirror and side window and wind shield, E: wind shield and engine hood, F: oblique line from top left to bottom right, wind shield, G: headlight, H: engine hood and air grille, I: front bumper and ground, J: oblique line from top left to bottom right, engine hood. Aeroplane -- A: nose pointing to the right, B: undercarriage, C: jet engine, mainly tube-shape, sometimes with highly turned ellipse/circle, D: body with small windows or text, E: wing or elevator tip pointing to the left, F: vertical stabilizer, with contour from top left to bottom right, G: vertical stabilizer base and body, with contour from top left to bottom right.}
\vspace{-0.5cm}
\label{fig:annotation}
\end{figure}

\section{ImageNetPart Dataset and Semantic Parts}
\label{sec:compart}
The problem of PASCAL3D+ dataset for evaluating visual concepts is that there are limited number of keypoint annotations, much fewer than the number of visual concepts, and they only cover a small amount of the object. This makes it hard to understand what all the visual concepts are doing. This motivated us to construct a new dataset, called ImageNetPart, which provides richer semantic part annotations using ImageNet images from PASCAL3D+. We focus on six vehicle categories: car, aeroplane, bicycle, motorbike, bus, and train. These annotations cover most of the objects and hence are more suitable for evaluating the visual concepts and understanding their correspondence to semantic parts.

Our annotation strategy is as follows. Firstly, we define semantic parts based on the object structure using terms like ``side mirror and side window and side body of car from left side". These semantic parts have some viewpoint information since the appearance of an object part may vary with viewpoint. Each defined semantic part represents one visual pattern. We manually construct the semantic part dictionary for each object class. Secondly, we ask labelers to locate each semantic part of specific category by putting a bounding box in the correct region of the image. To help the labelers understand the semantic parts, we provide them with two types of information during labeling: 1) one-sentence verbal description; 2) typical labeled bounding boxes on several prototype images with various viewpoints. Figure \ref{fig:annotation} visualizes annotations for several semantic parts, and we give the verbal descriptions for all the semantic parts of six object classes in the supplementary material.

\section{Experimental Results}\label{sec:exp}
\textbf{Dataset:} We use the PASCAL3D+ dataset \cite{xiang2014beyond}) and our newly annotated ImageNetPart dataset for quantitative evaluation. For PASCAL3D+ we experimented with six vehicle categories: car, aeroplane, bicycle, motorbike, bus, and train. ImageNetPart dataset provides much richer part annotations using ImageNet images in PASCAL3D+. For each category, we use the training subset for learning the visual concepts and the validation subset for testing.

The keypoint annotations from PASCAL3D+ are side-dependent, e.g., the left-front tires and the right-front tires have different labels. However the local image patches are not big enough to make such fine-grained distinctions possible. Therefore we merge some keypoint annotations.  The merging is done in the following cases. a) We merge the left/right fine-scale keypoints (e.g. left/right headlight) into a single class (e.g. headlight). b) We merge visually similar keypoints into a single class (e.g. the four wheels of a car). And, c) we discard keypoints that rarely occur in the images.

\textbf{Baselines:} We provide two baselines for comparison on PASCAL3D+. i) We evaluate keypoint detection performance for single filters. We use the filter responses, instead of distance to center of visual concept, as the detection scores when computing AP values. We report results for the best performing filters for each keypoint. ii) To provide an upper-bound on the detection capacity of unsupervised part-learning techniques, we report results obtained by using strong supervision. We extract the population responses for patches centered at the keypoint position (and randomly sampled patches for a separate background class) and train a multi-class linear SVM to discriminate the patches of different keypoints. Note that this baseline is very similar to the keypoint prediction method in \cite{long2014convnets} but with a different performance measure.

\textbf{Implementation Details}: In the experiments, we use the Caffe toolbox \cite{jia2014caffe} and VGG-16 network \cite{SimonyanZ14a} pre-trained on ImageNet object classification as our object-CNN. We build separate dictionaries for different object classes. For each image, we crop the object instance using the bounding box annotation and resize the short side to 224 pixels. We normalize each population response (feature vector) using $\ell_2$-normalization for both learning visual concepts and detection. We do not use $\ell_2$-normalization for evaluation of single filter method. Detailed comparison is in the diagnostic section.

\begin{table}[t!]
\centering
\tabcolsep=0.05cm
\begin{tabular}{ |c||c|c|c|c|c|c|c|c||c|c|c|c|c| }
\hline
        & \multicolumn{8}{|c||}{\textbf{Car}} & \multicolumn{5}{|c|}{\textbf{Motorbike}} \\
\hline
        & 1 & 2 & 3 & 4 & 5 & 6 & 7 & {\scriptsize mAP} & 1 & 2 & 3 & 4 & {\scriptsize mAP} \\
\hline
SF      & .86  &  .44  &  .30  &  .38  &  .19  &  .33   &  .13  & .38           &  .26  &  .60   &  .30  &  .22  & .35  \\
VC      & .92  &  .51  &  .27  &  .41  &  .36  &  .46   &  .18  & .45              &  .35  &  .75   &  .43  &  .25  & .45  \\
SS      & .97  &  .65  &  .37  &  .76  &  .45  &  .57   &  .30  & .58           &  .37  &  .77   &  .50  &  .60  & .56  \\
\hline
\end{tabular}

\vspace{0.3cm}
\begin{tabular}{ |c||c|c|c|c|c|c|c||c|c|c|c|c|c| }
\hline
       & \multicolumn{7}{|c||}{\textbf{Bus}} &  \multicolumn{6}{|c|}{\textbf{Bicycle}} \\
\hline
       & 1 & 2 & 3 & 4 & 5 & 6 & {\scriptsize mAP} & 1 & 2 & 3 & 4 & 5 & {\scriptsize mAP} \\
\hline
SF      & .45  &  .42   &  .23  &  .38  &  .80   &  .22  & .42      &  .23  &  .67  &  .25  &  .43  &  .43  &  .40  \\
VC   & .41  &  .59   &  .26  &  .29  &  .86   &  .51  & .49         &  .32  &  .78  &  .30  &  .55  &  .57  &  .50  \\
SS      & .74  &  .70   &  .52  &  .63  &  .90   &  .61  & .68      &  .37  &  .80  &  .34  &  .71  &  .64  &  .57  \\
\hline
\end{tabular}

\vspace{0.3cm}
\begin{tabular}{ |c||c|c|c|c|c|c||c|c|c|c|c|c| }

\hline
       & \multicolumn{6}{|c||}{\textbf{Train}} & \multicolumn{6}{|c|}{\textbf{Aeroplane}} \\
\hline
        & 1 & 2 & 3 & 4 & 5 & {\scriptsize mAP} & 1 & 2 & 3 & 4 & 5 & {\scriptsize mAP} \\
\hline
SF           & .39  &  .33   &  .24  &  .16  &  .15  & .25      & .41  &  .25  &  .22  &  .13  &  .31  &  .26  \\
VC           & .41  &  .30   &  .30  &  .28  &  .24  & .30      & .21  &  .47  &  .31  &  .16  &  .34  &  .30  \\
SS           & .71  &  .49   &  .50  &  .36  &  .39  & .49      & .72  &  .60  &  .50  &  .32  &  .49  &  .53  \\
\hline
\end{tabular}
\caption{\small AP values for keypoint detection on PASCAL3D+ dataset for six object categories using the ``best visual concept". ``SF" refers to single filter; ``VC" refers to visual concept; ``SS" refers to strong supervision. Visual concept achieves much higher AP than single filter method. The keypoint number-name mapping is provided below. Cars -- 1: wheel 2: wind shield 3: rear window 4: headlight 5: rear light 6: front 7: side; Bicycle --1: head center 2: wheel 3: handle 4: pedal 5: seat; Motorbike -- 1: head center 2: wheel 3: handle 4: seat; Bus -- 1: front upper corner 2: front lower corner 3: rear upper corner 4: rear lower corner 5: wheel 6: front center; Train -- 1: front upper corner 2: front lower corner 3: front center 4: upper side 5: lower side; Aeroplane -- 1: nose 2: upper rudder 3: lower rudder 4: tail 5: elevator and wing tip.}
\vspace{-0.5cm}
\label{tab:avg_pre}
\end{table}

\subsection{Results on PASCAL3D+}
We concentrate on the results using the visual concepts learnt from \texttt{pool4} with cluster merging described. We have 211, 237, 201, 176, 257 and 246 visual concepts for aeroplane, bicycle, bus, car, motorbike and train respectively. We show results from other layers in the diagnostic section. Note that here we report the results only for the images in PASCAL3D+ which lie in ImageNet subset (we obtained similar results for the much smaller subset of images which lie in PASCAL).

Table \ref{tab:avg_pre} reports the average precision (AP) values for keypoint detection in PASCAL3D+. For each keypoint from a particular object class, we show the AP achieved by the best visual concept and the best single filter (``best" meaning the one with highest AP for that keypoint). ``SF" refers to the baseline where we test single filters as keypoint detectors, and ``SS" refers to the baseline where we use strong supervision to train linear SVMs as keypoint detectors. The general trend shown in table \ref{tab:avg_pre} is that visual concepts are better as part detector than single filters. Unsurprisingly, the strongly supervised method is better than two unsupervised part-learning methods. Moreover, we see that the visual concepts obtained in our unsupervised way achieve reasonably good AP values for all categories.

\begin{table}[t]
\centering
\tabcolsep=0.06cm
\begin{tabular}{ |c||c|c|c|c|c|c|c|c|c|c|c|c|c| }
\hline
\textbf{Car} & 1 & 2 & 3 & 4 & 5 & 6 & 7 & 8 & 9 & 10 & 11 & 12 \\
\hline
SF      & .86  &  .87  &  .84  &  .68  &  .70  &  .83  &  .83   & .82  & .72  &  .18  &  .22  &  .49   \\
VC       & .94  &  .97  &  .94  &  .93  &  .94  &  .95  &  .94   & .92  & .94  &  .42  &  .48  &  .58   \\
\hline \hline
\textbf{Car} & 13 & 14 & 15 & 16 & 17 & 18 & 19 & 20 & 21 & 22 & 23 & 24  \\
\hline
SF &  .33  &  .25  &  .22  &  .18   &  .37  &  .18  & .28  &  .13  &  .16  &  .14  &  .30  &  .20    \\
VC  &  .45  &  .48  &  .46  &  .54   &  .60  &  .34  & .38  &  .19  &  .26  &  .37  &  .42  &  .32    \\
\hline \hline
\textbf{Car} & 25 & 26 & 27 & 28 & 29 & 30 & 31 & 32 & 33 & 34 & 35 & {\scriptsize mAP} \\
\hline
SF      &  .16   & .17  & .15  &  .27  &  .19  &  .25  &  .18  &  .19  &  .16  &  .07   &  .07  &  .36 \\
VC       &  .26   & .40  & .29  &  .40  &  .18  &  .41  &  .53  &  .31  &  .57  &  .36   &  .26  &  .53 \\
\hline
\end{tabular}

\vspace{0.2cm}
\begin{tabular}{ |c||c|c|c|c|c|c|c|c|c|c| }
\hline
\textbf{Aeroplane} & 1 & 2 & 3 & 4 & 5 & 6 & 7 & 8 & 9  \\
\hline
SF      & .32  &  .42  &  .08  &  .61  &  .58  &  .42  &  .49   & .31  & .23       \\
VC       & .26  &  .21  &  .07  &  .84  &  .61  &  .44  &  .63   & .42  & .34      \\
\hline \hline
\textbf{Aeroplane} & 10 & 11 & 12 & 13 & 14 & 15 & 16 & 17 & {\scriptsize mAP}  \\
\hline
SF   &  .09  &  .07  &  .20   &  .12  &  .29  &  .29  &  .07   &  .09  &  .28   \\
VC    &  .15  &  .11  &  .44   &  .23  &  .59  &  .65  &  .08   &  .15  &  .37   \\
\hline
\end{tabular}

\vspace{0.2cm}
\begin{tabular}{ |c||c|c|c|c|c|c|c|c|c|c|c|c|c|c|c|c|c|c|c| }
\hline
\textbf{\small{Bike}} & 1 & 2 & 3 & 4 & 5 & 6 & 7 & 8 & 9 & 10 & 11 & 12 & 13 & {\scriptsize mAP} \\
\hline
\small
\small{SF}      &  \small{.77}  &  \small{.84}   & \small{.89}  & \small{.91}  &  \small{.94}  &  \small{.92}  &  \small{.94}  &  \small{.91}  &  \small{.91}  &  \small{.56}  &  \small{.53}   &  \small{.15}  &  \small{.40}  &  \small{.75}  \\
\small{VC}   &  \small{.91}  &  \small{.95}   & \small{.98}  & \small{.96}  &  \small{.96}  &  \small{.96}  &  \small{.97}  &  \small{.96}  &  \small{.97}  &  \small{.73}  &  \small{.69}   &  \small{.19}  &  \small{.50}  &  \small{.83}  \\
\hline
\end{tabular}

\vspace{0.2cm}
\begin{tabular}{ |c||c|c|c|c|c|c|c|c|c|c|c|c|c|c|c|c|c|c|c| }
\hline
\textbf{Motor} & 1 & 2 & 3 & 4 & 5 & 6 & 7 & 8 & 9 & 10 & 11 & 12 & {\scriptsize mAP} \\
\hline
SF      &  .69  &  .46  &  .76   & .67  & .66  &  .57  &  .54  &  .70  &  .68  &  .25  &  .17  &  .22   &  .53   \\
VC   &  .89  &  .64  &  .89   & .77  & .82  &  .63  &  .73  &  .75  &  .88  &  .39  &  .33  &  .29   &  .67   \\
\hline
\end{tabular}

\vspace{0.2cm}
\begin{tabular}{ |c||c|c|c|c|c|c|c|c|c|c|c|c|c|c|c|c|c|c|c| }
\hline
\textbf{Bus} & 1 & 2 & 3 & 4 & 5 & 6 & 7 & 8 & 9 & {\scriptsize mAP} \\
\hline
SF      & .90  & .40  &  .49  &  .46  &  .31  &  .28  &  .36  &  .38   &  .31  &  .43  \\
VC   & .93  & .64  &  .69  &  .59  &  .42  &  .48  &  .39  &  .32   &  .27  &  .53  \\
\hline
\end{tabular}

\vspace{0.2cm}
\begin{tabular}{ |c||c|c|c|c|c|c|c|c|c|c|c|c|c|c|c|c|c|c|c| }
\hline
\textbf{Train} & 1 & 2 & 3 & 4 & 5 & 6 & 7 & 8 \\
\hline
SF      &  .58  &  .07  &  .20  &  .15   & .21  & .15  &  .27  & .43 \\
VC   &  .66  &  .50  &  .32  &  .28   & .24  & .15  &  .33  & .72 \\
\hline
\textbf{Train} & 9 & 10 & 11 & 12 & 13 & 14 & {\scriptsize mAP} & \\
\hline
SF  &  .17  &  .27  &  .16  &  .25  &  .17   &  .10  &  .23 & \\
VC  &  .36  &  .41  &  .27  &  .45  &  .27   &  .47  &  .39 & \\
\hline
\end{tabular}

\caption{\small AP values for keypoint detection on ImageNetPart dataset for cars (top), aeroplanes (middle) and bicycle (bottom). We see that visual concept by population encoding achieves much higher AP than single filter method. The semantic part names are provided in the supplementary material.}
\label{tab:compart_PR}
\vspace{-0.5cm}
\end{table}

\subsection{Results on ImageNetPart}
The richer annotations on ImageNetPart allows us to do more detailed experiments which make it much clearer how the visual concepts correspond to the object semantic parts. We start by evaluating the ``best visual concepts" for each semantic part following the strategy discussed above for PASCAL3D+. The results are fairly similar showing that visual concepts can be very effective at detecting most semantic parts and generally perform better than single filters. From table \ref{tab:compart_PR} we observe the same trends as for the keypoint results, the ``best visual concept" results are best on cars, bicycles, motorbikes, and buses.

\begin{table}[t]
\centering
\tabcolsep=0.05cm
\begin{tabular}{|l || c|c|c|c|c|c|c|c|}
\hline
   & 1 & 2 & 3 & 4 & 5 & 6 & 7 & {\scriptsize mAP} \\
\hline
VC-512     &  .94   &  .51  &  .32  &  .53  &  .36  &  .48   &  .22  & .48 \\
VC-256     &  .94   &  .47  &  .29  &  .54  &  .31  &  .45   &  .20  & .45 \\
VC-128     &  .94   &  .46  &  .27  &  .51  &  .17  &  .47   &  .19  & .43 \\
VC-64      & .93    &  .47  &  .27  &  .29  & .17   &  .46   &  .18  & .40 \\
Merge   & .92    &  .51  &  .27  &  .41  & .36   &  .46   &  .18  & .45 \\
SF         & .86    &  .44  &  .30  &  .38  &  .19  &  .33   &  .13  & .38 \\
\hline
\end{tabular}
\caption{\small The effect of $K$ on keypoint detection performance. "VC-$K$" refers to setting $K$ as the number of clusters in kmeans++. All the VC-based methods are without merging.}
\label{tab:AP_K}

\centering
\tabcolsep=0.05cm
\begin{tabular}{|l || c|c|c|c|c|c|c|c|}
\hline
                & 1 & 2 & 3 & 4 & 5 & 6 & 7 & {\scriptsize mAP} \\
\hline
VC-NL2      &  .92   &  .33  &  .17  &  .34   &  .12  &  .39   &  .16  & .35  \\
VC-L2       &  .94   &  .51  &  .32  &  .53   &  .36  &  .48   &  .22  & .48  \\
SF-NL2   &  .86   &  .44  &  .30  &  .38   &  .19  &  .33   &  .13  & .38  \\
SF-L2    &  .80   &  .42  &  .24  &  .42   &  .19  &  .31   &  .18  & .37  \\
\hline
\end{tabular}
\caption{\small The effect of $\ell_2$-normalization for cluster and single filter. ``NL2" refers to not using $\ell_2$-normalization, and ``L2" refers to using $\ell_2$-normalization. We can see that $\ell_2$-normalization is crucial to population encoding.}
\label{tab:L2_norm}

\centering
\tabcolsep=0.05cm
\begin{tabular}{|l || c|c|c|c|c|c|c|c|}
\hline
Layer               & 1 & 2 & 3 & 4 & 5 & 6 & 7 & {\scriptsize mAP} \\
\hline
pool3      & .80  &  .28  &  .13  &  .22  &  .26  &  .22  &  .16  & .29  \\
pool4      & .92  &  .51  &  .27  &  .41  &  .36  &  .46  &  .18  & .45  \\
pool5      & .83  &  .61  &  .27  &  .55  &  .29  &  .45  &  .29  & .47  \\
\hline
\end{tabular}
\caption{\small AP values obtained using visual concepts learnt from different layers for cars. Different layers of the CNN are capturing different scales of object parts. \texttt{pool5} is better on ``big" parts while \texttt{pool4} is better on ``small" parts.}
\label{tab:car_other_layers}

\centering
\tabcolsep=0.05cm
\begin{tabular}{|l || c|c|c|c|c|c|c|c|}
\hline
Method                & 1 & 2 & 3 & 4 & 5 & 6 & 7 & {\scriptsize mAP} \\
\hline
SF        & .96   &  .65   &  .61   &  .59   &  .59  &  .38   &  .27  & .58  \\
VC      & .99   &  .75   &  .61   &  .73   &  .70  &  .54   &  .39  & .67  \\
SS        & .99   &  .87   &  .82   &  .88   &  .84  &  .62   &  .41  & .78  \\
\hline
\end{tabular}
\caption{\small AP values of viewpoint control for cars. The results are higher than the car results in Table 1 of the paper after selecting the best viewpoint for each keypoint.\newline}
\label{tab:car_vp}
\vspace{-0.5cm}
\end{table}

\subsection{Diagnostic Experiments}
\label{sec:diag}
In this section we present results from some diagnostic experiments to enable a better understanding of the visual concepts obtained from population encoding. Currently, our experiments are performed on cars using PASCAL3D+ dataset. Experiments on other object categories or other dataset is left as future work.

\textbf{Effect of the number of clusters $K$:}
An important issue is to determine a good value for the number of clusters $K$. This is partly influenced by the amount of data available because we risk over-fitting if we try to make $K$ too large. But it is also affected by the variability in appearance and geometry of the object. The greater the variability of the object, presumably, the larger the number of visual concepts. But it is unclear how to measure the variability of the object in advance. Determining the number $K$ of clusters is important in order to prevent over-fitting while capturing the variability of the feature responses. We resort to a two-pronged strategy: (i) we choose a range of values for $K$ (64, 128, 256, 512) to see how the results change with $K$, and (ii) we use a second stage to reduce the number of clusters by a merging algorithm.

Our results, Table \ref{tab:AP_K}, show that although performance increases with $K$, the improvement is relatively slow from $K=64$ to $K=512$. Note that even at $K=64$ the performance by visual concept is still better than single filter (``SF") method. Also the merging algorithm (``Merge") achieves the same AP as VC-256, but with much fewer visual concepts (176).

\textbf{Effect of $\ell_2$-normalization:} We make a comparison between using $\ell_2$-normalization as a preprocessing step and without $\ell_2$-normalization. From Table \ref{tab:L2_norm}, we can see that $\ell_2$-normalization is critical for visual concepts obtained from population encoding method. We then hypothesize that it is the population response direction that captures visual information, instead of the population response magnitude. For the single filter case, we have also tried similar $\ell_2$-normalization across filters for each spatial position, and used the normalized response as detection score. But we find that $\ell_2$-normalization in single filter case makes little difference (slightly worse) in terms of the AP values.

\textbf{Effect of Intermediate Layers:} In Table \ref{tab:car_other_layers}, we present the results obtained using the visual concepts learnt from different layers (\texttt{pool3} - \texttt{pool5}). Looking at the numbers from \texttt{pool3}, it is clear that its concept dictionaries are not able to capture semantic level of the part annotations. This is because the receptive field of \texttt{pool3} is too small to capture the entire part. Surprisingly, \texttt{pool5} gives similar results as pool4 layer. But on careful examination, we see that \texttt{pool5} does better than \texttt{pool4} when the keypoints correspond to ``big" parts (e.g. windshield) and worse than \texttt{pool4} when the keypoints correspond to smaller parts (e.g. wheels), the size of which is more appropriate for \texttt{pool4}. This shows that different layers of the CNN are capturing different scales of object parts.

\textbf{Viewpoint Control:} While visualizing the clusters, we find that many visual concepts are viewpoint specific. Therefore, we test the performance of visual concepts by controlling for viewpoint. We divide the test set into five viewpoint bins: front, front-side, side, rear-side, and rear. Then for each viewpoint we run the same evaluation procedure as described above. Table \ref{tab:car_vp} shows the best viewpoint AP result for each keypoint. The numbers clearly shows a significant jump in detection capabilities of both single filters and our visual concepts compared to the car results in Table 1 in the main paper. Note that population encoding still outperforms single-filter detectors. These results suggest the lack of viewpoint invariance in the internal representation of the CNNs.

\begin{figure}[h!]
\centering
\subfloat[car]
{\includegraphics[width=.45\linewidth, height=.25\linewidth]{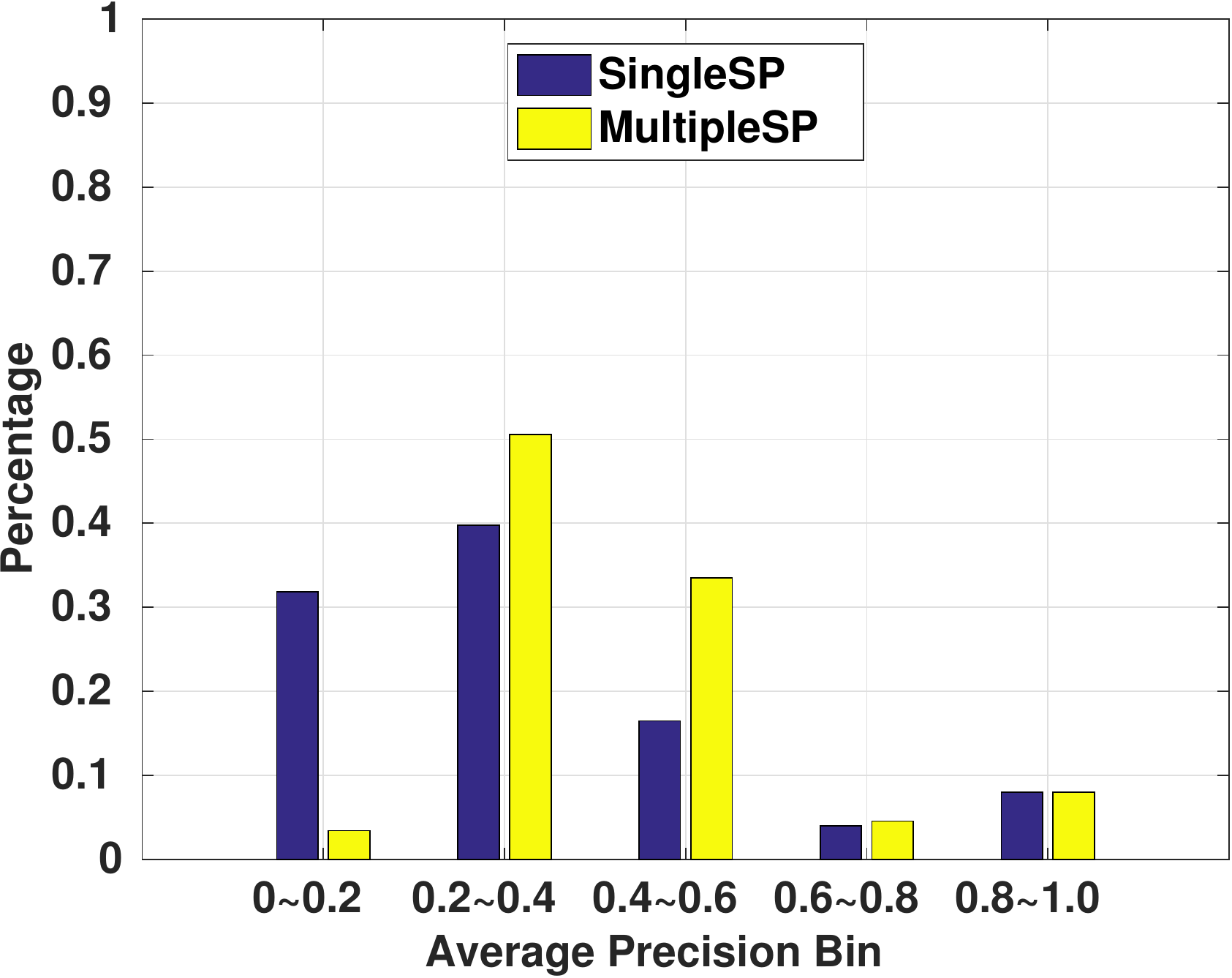}}
\hspace{0.3cm}
\subfloat[aeroplane]
{\includegraphics[width=.45\linewidth, height=.25\linewidth]{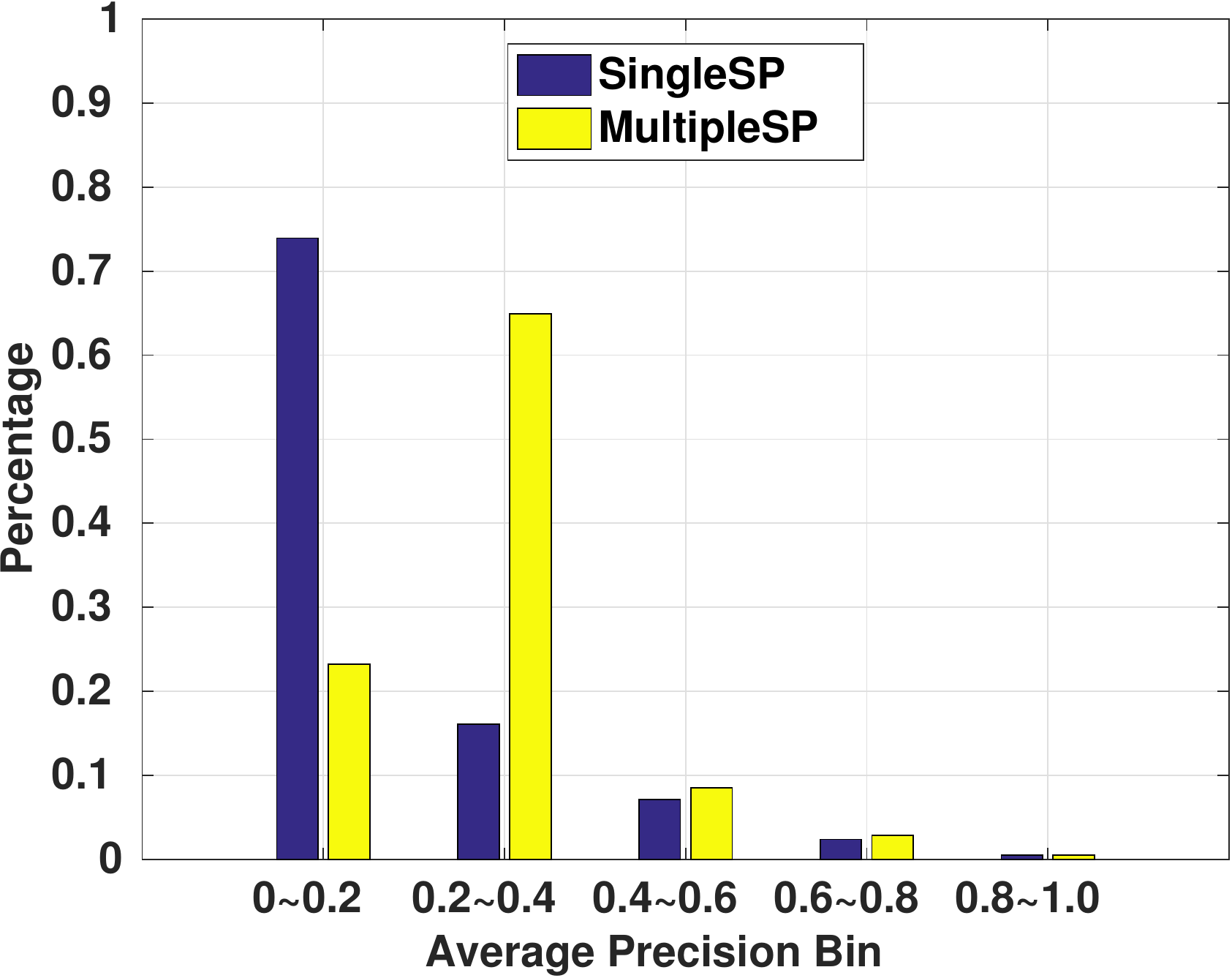}}
\vspace{-0.2cm}

\subfloat[bus]
{\includegraphics[width=.45\linewidth, height=.25\linewidth]{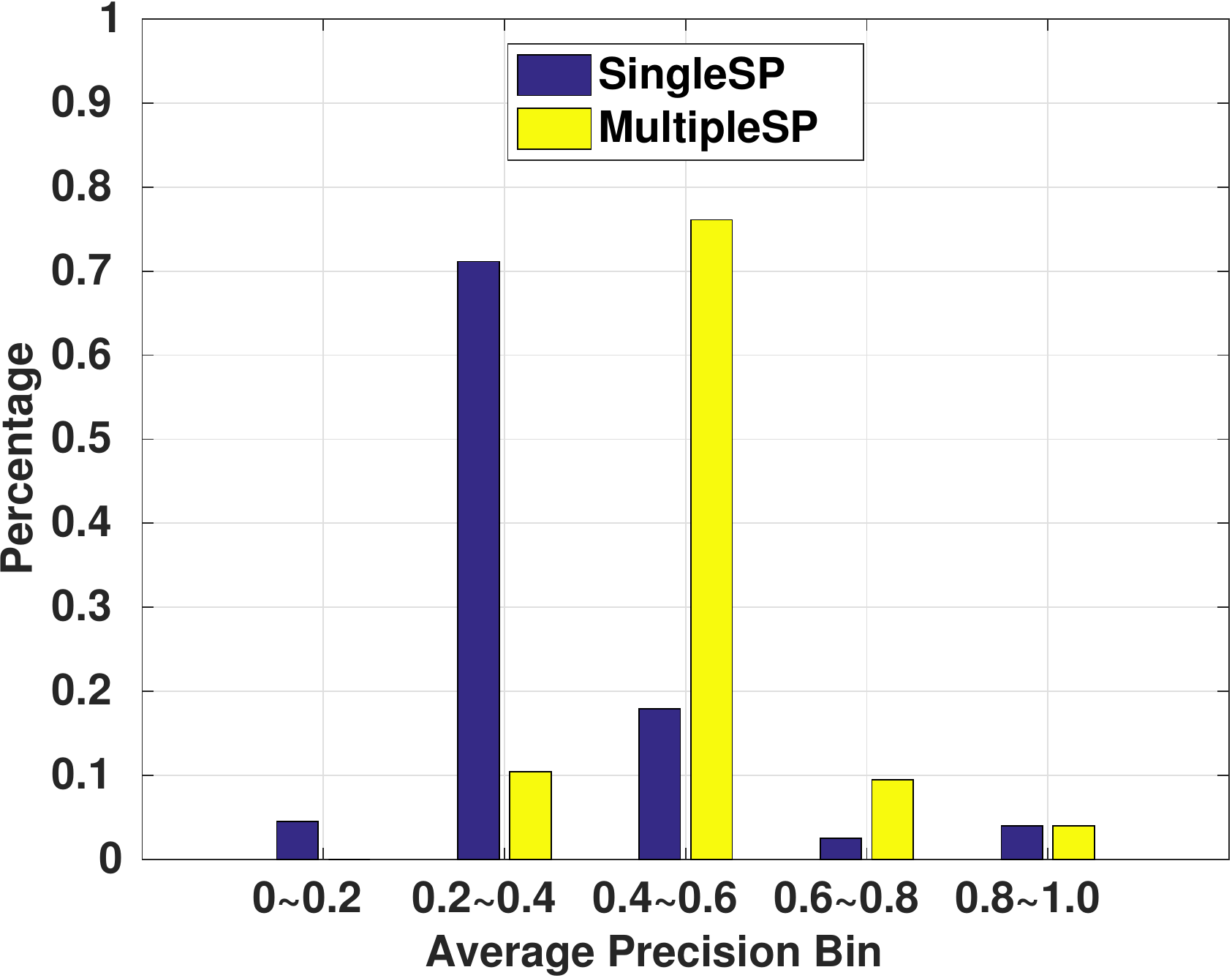}}
\hspace{0.3cm}
\subfloat[train]
{\includegraphics[width=.45\linewidth, height=.25\linewidth]{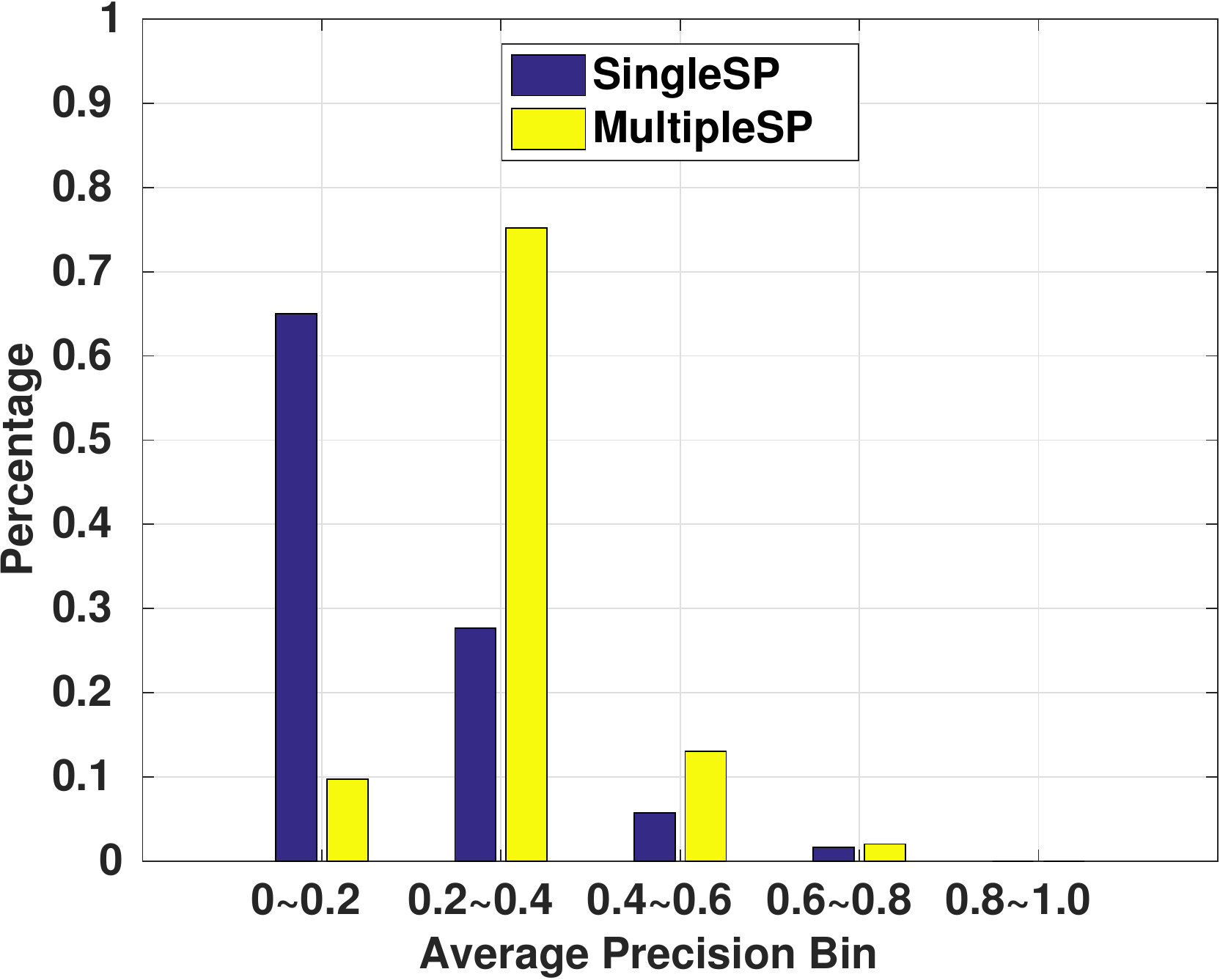}}

\vspace{-0.2cm}
\subfloat[bicycle]
{\includegraphics[width=.45\linewidth, height=.25\linewidth]{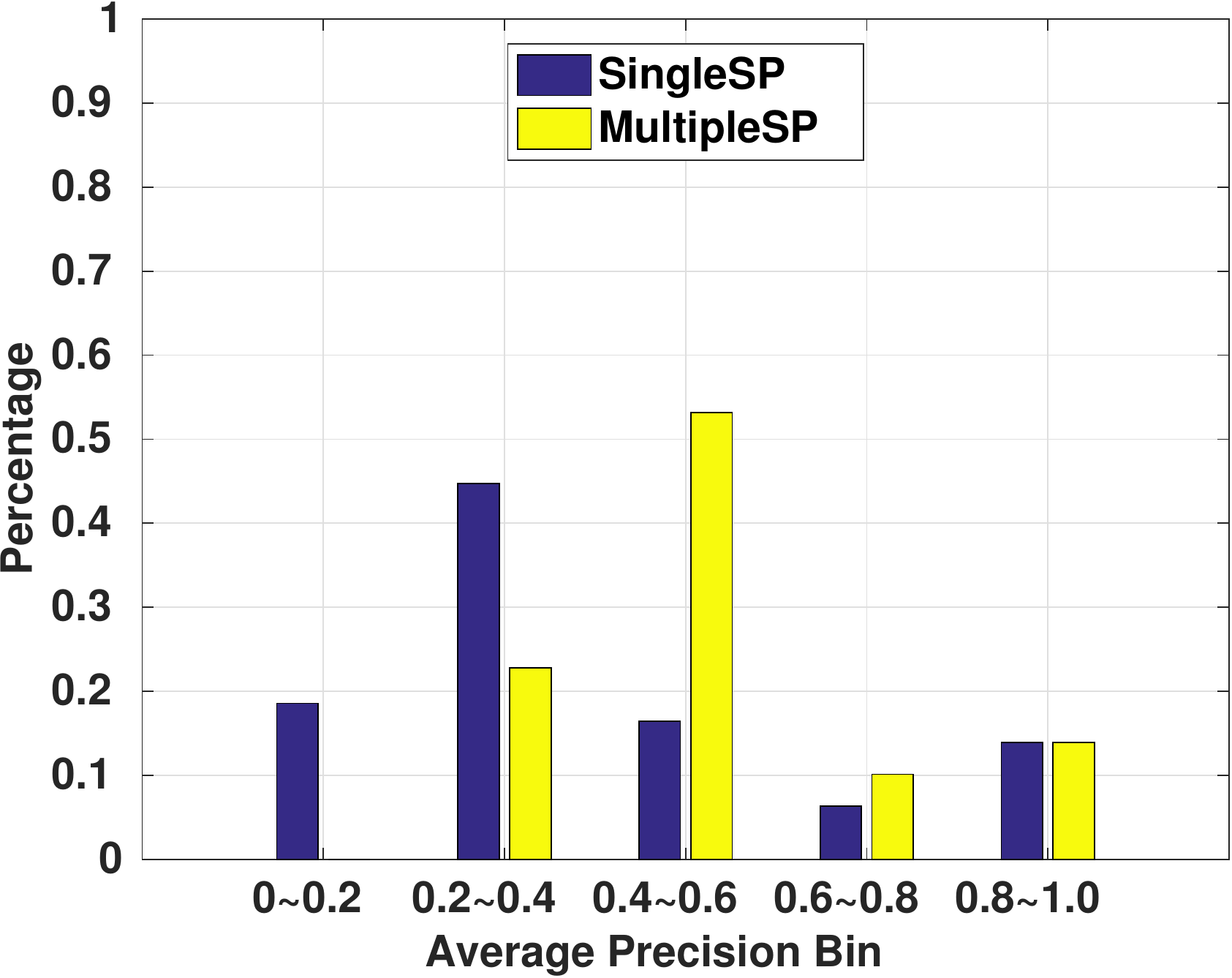}}
\hspace{0.3cm}
\subfloat[motorbike]
{\includegraphics[width=.45\linewidth, height=.25\linewidth]{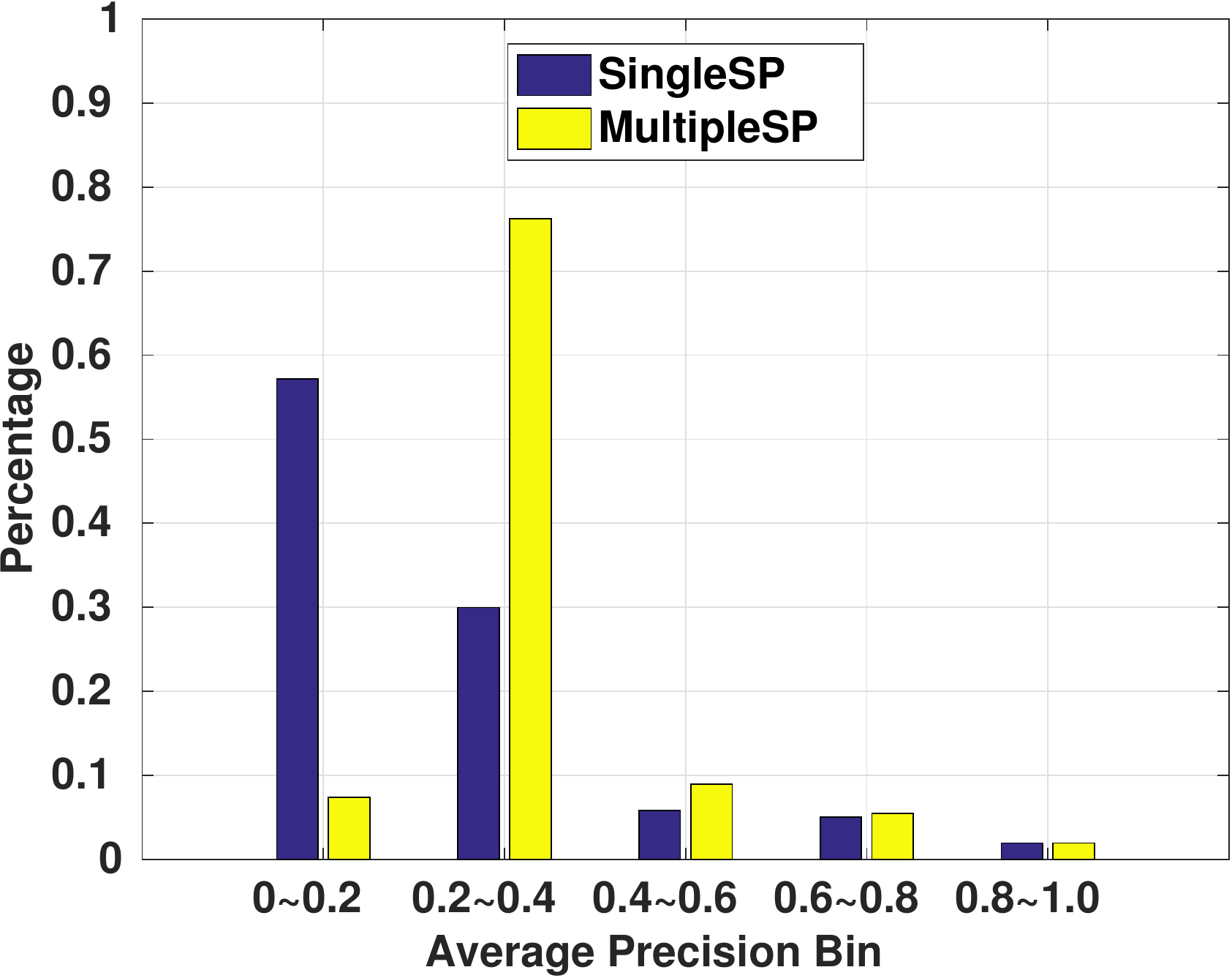}}
\caption{\small The histograms of the AP responses for all visual concepts. For ``SingleSP", we evaluate each visual concept by reporting its AP for its best semantic part (the one it best detects). Some visual concepts have very low APs when evaluated in this manner. For ``MultipleSP", we allow each visual concept to detect a small subset of semantic parts (two, three, or four) and report the AP for the best subset (note, the visual concept is penalized if it does not detect all of them). The APs rise considerably using the MultipleSP evaluation, suggesting that some of the visual concepts detect a subset of semantic parts. The remaining visual concepts with very low APs correspond to the background.}
\label{fig:many}
\end{figure}

\begin{figure}[t!]
\centering
\includegraphics[width=\linewidth, height=.55\linewidth]{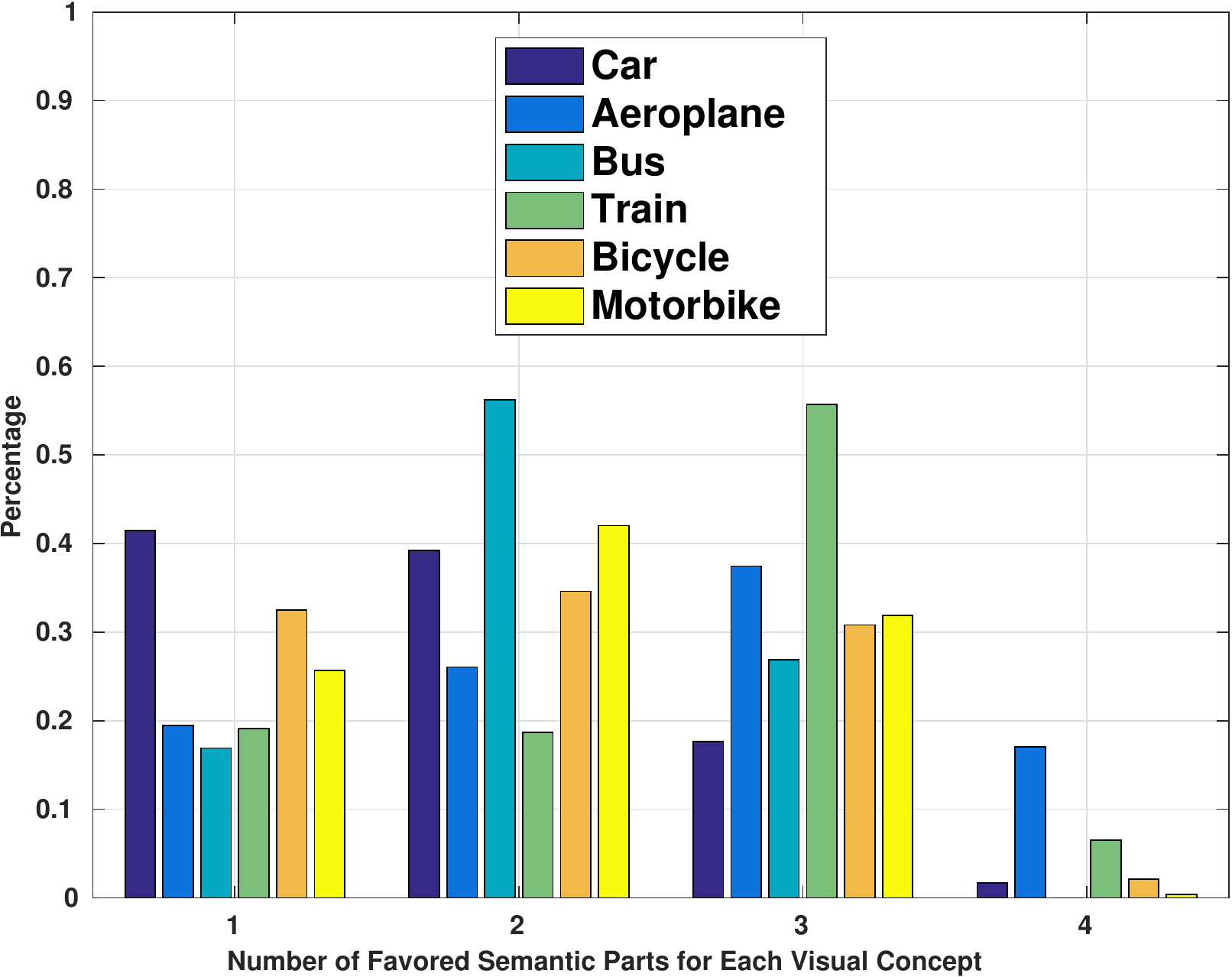}
\caption{\small Distribution of number of semantic parts favored by each visual concept for 6 objects. Most visual concepts like one, two, or three semantic parts.}
\label{fig:sp_num}
\vspace{-0.5cm}
\end{figure}

\subsection{The correspondence between Visual Concepts and Semantic Parts}
\begin{figure}[t!]
  \centering
\begin{minipage}{.32\textwidth}
  \includegraphics[width=0.85cm,height=0.85cm]{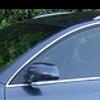}
  \includegraphics[width=0.85cm,height=0.85cm]{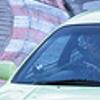}
  \includegraphics[width=0.85cm,height=0.85cm]{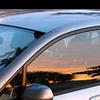}
  \includegraphics[width=0.85cm,height=0.85cm]{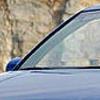}
  \includegraphics[width=0.85cm,height=0.85cm]{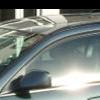}
  \includegraphics[width=0.85cm,height=0.85cm]{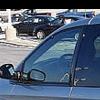}
\end{minipage}
\hfill
\begin{minipage}{.055\textwidth}
  \includegraphics[width=0.85cm,height=0.85cm]{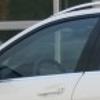}
\end{minipage}
\hfill
\begin{minipage}{.055\textwidth}
  \includegraphics[width=0.85cm,height=0.85cm]{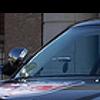}
\end{minipage} \\ \vspace{.2cm}
\begin{minipage}{.32\textwidth}
  \includegraphics[width=0.85cm,height=0.85cm]{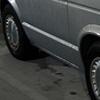}
  \includegraphics[width=0.85cm,height=0.85cm]{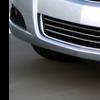}
  \includegraphics[width=0.85cm,height=0.85cm]{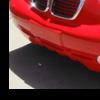}
  \includegraphics[width=0.85cm,height=0.85cm]{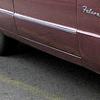}
  \includegraphics[width=0.85cm,height=0.85cm]{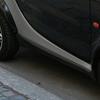}
  \includegraphics[width=0.85cm,height=0.85cm]{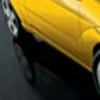}
\end{minipage}
\hfill
\begin{minipage}{.055\textwidth}
  \includegraphics[width=0.85cm,height=0.85cm]{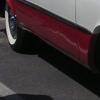}
\end{minipage}
\hfill
\begin{minipage}{.055\textwidth}
  \includegraphics[width=0.85cm,height=0.85cm]{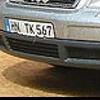}
\end{minipage}
\caption{\small This figure shows two examples of visual concepts that correspond to multiple semantic parts. The first six columns show example patches from that visual concept, and the last two columns show prototypes from two different corresponding semantic parts. We see that the visual concept in the first row corresponds to ``side window" and ``front window", and the visual concept in the second row corresponds to ``side body and ground" and ``front bumper and ground". Note that these semantic parts look fairly similar.}
\label{fig:dual_SP}
\end{figure}

\begin{figure}[t!]
  \centering
\begin{minipage}{.5\textwidth}
  \includegraphics[width=0.96cm,height=0.96cm]{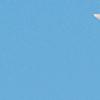}
  \includegraphics[width=0.96cm,height=0.96cm]{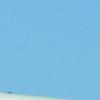}
  \includegraphics[width=0.96cm,height=0.96cm]{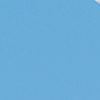}
  \includegraphics[width=0.96cm,height=0.96cm]{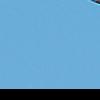}
  \includegraphics[width=0.96cm,height=0.96cm]{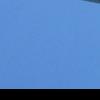}
  \includegraphics[width=0.96cm,height=0.96cm]{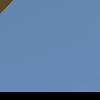}
  \includegraphics[width=0.96cm,height=0.96cm]{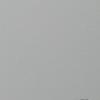}
  \includegraphics[width=0.96cm,height=0.96cm]{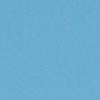}
\end{minipage} \\ \vspace{.2cm}
\begin{minipage}{.5\textwidth}
  \includegraphics[width=0.96cm,height=0.96cm]{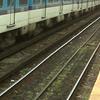}
  \includegraphics[width=0.96cm,height=0.96cm]{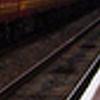}
  \includegraphics[width=0.96cm,height=0.96cm]{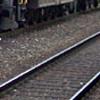}
  \includegraphics[width=0.96cm,height=0.96cm]{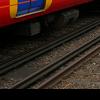}
  \includegraphics[width=0.96cm,height=0.96cm]{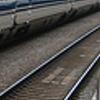}
  \includegraphics[width=0.96cm,height=0.96cm]{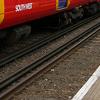}
  \includegraphics[width=0.96cm,height=0.96cm]{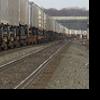}
  \includegraphics[width=0.96cm,height=0.96cm]{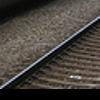}
\end{minipage} \\ \vspace{.2cm}
\begin{minipage}{.5\textwidth}
  \includegraphics[width=0.96cm,height=0.96cm]{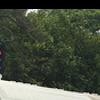}
  \includegraphics[width=0.96cm,height=0.96cm]{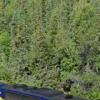}
  \includegraphics[width=0.96cm,height=0.96cm]{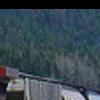}
  \includegraphics[width=0.96cm,height=0.96cm]{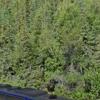}
  \includegraphics[width=0.96cm,height=0.96cm]{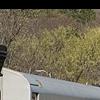}
  \includegraphics[width=0.96cm,height=0.96cm]{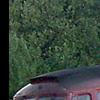}
  \includegraphics[width=0.96cm,height=0.96cm]{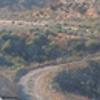}
  \includegraphics[width=0.96cm,height=0.96cm]{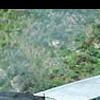}
\end{minipage}
\caption{\small The first row shows a visual concept corresponding to sky from the aeroplane class, and the second and third rows are visual concepts corresponding to railway track and tree respectively from the train class.}
\label{fig:background}
\vspace{-0.5cm}
\end{figure}

The ``best visual concept" evaluation strategy is problematic because it assumes that each visual concept detects a single semantic part only. Moreover, the number of visual concepts is much larger than the number of semantic parts so it does not help understand what the majority of the visual concepts are doing. Hence for each visual concept we analyze how well it responds to one semantic part, or a small subset. Firstly, for each visual concept we determine which semantic part it best detects, calculate the AP, and plot the histogram as shown in figure \ref{fig:many} (SingleSP). Next, we allow each visual concept to select a small subset -- two, three, or four -- of semantic parts that it can detect (penalizing it if it fails to detect all of them). We measure how well the visual concept detects this subset using AP and plot the same histogram as before. This generally shows much better performance than before. From figure \ref{fig:many} (MultipleSP) we see that the histogram of APs gets shifted greatly to the right, when taking into account the fact that one visual concept may correspond to one or more semantic parts. Figure \ref{fig:sp_num} shows the percentage of how many semantic parts are favored by each visual concept. If a visual concept responds well to two, or more, semantic parts, this is often because those parts are visually similar, see figure \ref{fig:dual_SP}. For some object classes, particularly aeroplanes and trains, there remain some visual concepts with low APs even after allowing multiple semantic parts. Our analysis shows that many of these visual concepts are detecting background (e.g., the sky for the aeroplane class, and railway tracks or coaches for the train class), see figure \ref{fig:background}.
A few others have no obvious interpretation and are probably due to limitations of the clustering algorithm and CNN feature.


Note that our results also imply that there are several visual concepts for each semantic part. In other work, in preparation, we show that these visual concepts typically correspond to different subregions of the semantic parts (with some overlap) and that combining them yields better detectors for all semantic parts with a mean gain of 0.25 AP.

\section{Conclusion and Future Work}
\label{sec:conclusion}

This paper hypothesizes that CNNs have internal representations of object parts which are encoded by population activity of featres/neurons. To validate this hypothesis we used clustering to determine a set of visual concepts for each object class at each layer of the CNN. We showed that these visual concepts are visually tight and appear to correspond to semantic parts of the object. This is supported by quantitative analysis which finds the ``best visual concept" for each keypoint (PASCAL3D+) and shows that they have high AP for detecting the keypoints. To further understand the relationship between visual concepts and semantic parts we constructed a new dataset ImageNetPart by providing dense annotations in terms of semantic parts. This dataset showed good results for the ``best visual concept" but also enabled us to show that most visual concepts were fairly effective at detecting a small subset of semantic parts with similar visual appearance (ranging from one to four). We conclude that visual concepts help understand the internal representations of deep networks used for object detection and that our clustering method for discovering visual concepts can be thought of as unsupervised learning of object parts. Future work involves building on ideas in  \cite{zhu2010learning} to learn  models, by grouping visual concepts which co-occur.

\newpage

\bibliography{egbib}
\bibliographystyle{ieee}
\end{document}